\documentclass[journal,compsoc,cspaper,final]{IEEEtran}
\usepackage{algorithmic}
\usepackage{algorithm}
\usepackage{array}
\usepackage[caption=false,font=normalsize,labelfont=sf,textfont=sf]{subfig}
\usepackage{url}
\usepackage{graphicx}
\usepackage{amsmath}
\usepackage{amssymb}
\usepackage{booktabs}
\usepackage[pagebackref,breaklinks,colorlinks]{hyperref}

\usepackage{times}
\usepackage{epsfig}
\usepackage[accsupp]{axessibility}  % Improves PDF readability for those with disabilities.
\usepackage{blindtext}
\usepackage{amsfonts}
\usepackage[inline]{enumitem}
\usepackage{algorithmic}
\usepackage{textcomp}

\usepackage{multirow}
\usepackage{float}
\usepackage{adjustbox}
\usepackage{etoolbox}
\usepackage{dsfont}
\usepackage{float}
\usepackage{lipsum}
\usepackage{stfloats}
\usepackage{multicol}
\usepackage{amsmath,bm}
\usepackage{array}
\usepackage{tabulary}
\usepackage{paralist}
\usepackage{bbding}
\usepackage{pifont}
\newcommand{\cmark}{\ding{52}}%
\newcommand{\xmark}{\ding{56}}%
\usepackage[table]{xcolor}
\usepackage{colortbl}
\usepackage{diagbox}
\usepackage[normalem]{ulem}
\definecolor{tableHeadGray}{gray}{.9}
\definecolor{tableSubHeadGray}{gray}{0.95}
\newcommand{\etal}{\textit{et al.}}
\newcommand{\PAR}[1]{\noindent{\bf #1}}
\newcommand*{\ourmodel}{REFID\@\xspace}
\newcommand*{\ourfusionmodule}{EGACA\@\xspace}
\newcommand*{\ourdataset}{HighREV\@\xspace}

\makeatletter
\DeclareRobustCommand\onedot{\futurelet\@let@token\@onedot}
\def\@onedot{\ifx\@let@token.\else.\null\fi\xspace}

\def\ie{\emph{i.e}\onedot} 
 
\def\etc{\emph{etc}\onedot} 
 
\def\etal{\emph{et al}\onedot}
\makeatother

\newcommand{\Add}[1]{\textcolor{black}{\textbf{}#1\textbf{}}}
\newcommand{\revise}[1]{\textcolor{black}{\textbf{}#1\textbf{}}}
\newcommand{\red}[1]{\textcolor{red}{#1}}

\ifCLASSOPTIONcompsoc
  \usepackage[nocompress]{cite}
\else
  \usepackage{cite}
\fi
\ifCLASSINFOpdf
\else
\fi
\begin{document}
\title{A Unified Framework for Event-based Frame Interpolation with Ad-hoc Deblurring in the Wild}

\author{Lei Sun, Daniel Gehrig, Christos Sakaridis, Mathias Gehrig, Jingyun Liang, Peng Sun, Zhijie Xu,\\Kaiwei Wang, Luc Van Gool, and Davide Scaramuzza% <-this % stops a space
\IEEEcompsocitemizethanks{\IEEEcompsocthanksitem L. Sun is with the National Research Center for Optical Instrumentation, Zhejiang University, 310027 Hangzhou, China, the Robotics and Perception Group, University of Zurich, 8050 Zurich, Switzerland.\protect\\
E-mail: leo\_sun@zju.edu.cn.

\IEEEcompsocthanksitem D. Gehrig, M. Gehrig, and D. Scaramuzza are with the Robotics and Perception Group, University of Zurich, 8050 Zurich, Switzerland.\protect\\
E-mails: \{dgehrig, mgehrig, sdavide\}@ifi.uzh.ch.

\IEEEcompsocthanksitem C. Sakaridis and J. Liang are with the Computer Vision Lab, ETH Zurich, 8092 Zurich, Switzerland.\protect\\
E-mails: \{csakaridis, jinliang\}@vision.ee.ethz.ch.

% \protect\\
% E-mails: \{lei.sun, vangool\}@vision.ee.ethz.ch.

\IEEEcompsocthanksitem P. Sun and K. Wang are with the  National Research Center for Optical Instrumentation, Zhejiang University, 310027 Hangzhou, China.\protect\\
E-mails: \{pengsunr, wangkaiwei\}@zju.edu.cn.

\IEEEcompsocthanksitem Zhijie Xu is with the Centre for Visual and Immersive Computing, Huddersfield University, HD1 3DH UK.\protect\\
E-mail: z.xu@hud.ac.uk.

\IEEEcompsocthanksitem L. Van Gool is with the INSAIT, Sofia University ``St. Kliment Ohridski'', Bulgaria.\protect\\
E-mail: vangool@vision.ee.ethz.ch.

\IEEEcompsocthanksitem This work was supported
in part by the European Research Council (ERC) under grant agreement No. 864042 (AGILEFLIGHT),
in part by the National Natural Science Foundation of China (NSFC) under Grant No.12174341, in part by the National Key R\&D Program of China under Grant No.2022YFF0705500 and No.2022YFB3206000, in part by the China Scholarship Council, and in part by AlpsenTek GmbH. \IEEEcompsocthanksitem Corresponding author: K. Wang.
}% <-this % stops an unwanted space

% \thanks{}
}


\markboth{IEEE TRANSACTIONS ON PATTERN ANALYSIS AND MACHINE INTELLIGENCE}%
{Shell \MakeLowercase{\textit{et al.}}: Bare Demo of IEEEtran.cls for Computer Society Journals}



\IEEEtitleabstractindextext{%
\begin{abstract}
% The performance of video frame interpolation is inherently correlated with the ability to handle motion in the input scene. Even though previous works recognize the utility of asynchronous event information for this task, they ignore the fact that motion may or may not result in blur in the input video to be interpolated, restricting themselves to frame interpolation. 
\Add{Effective video frame interpolation hinges on the adept handling of motion in the input scene. Prior work acknowledges asynchronous event information for this, but often overlooks whether motion induces blur in the video, limiting its scope to sharp frame interpolation. We instead propose a unified framework for event-based frame interpolation that performs deblurring ad-hoc and thus works both on sharp and blurry input videos. Our model consists in a bidirectional recurrent network that incorporates the temporal dimension of interpolation and fuses information from the input frames and the events adaptively based on their temporal proximity. To enhance the generalization from synthetic data to real event cameras, we integrate self-supervised framework with the proposed model to enhance the generalization on real-world datasets in the wild. At the dataset level, we introduce a novel real-world high-resolution dataset with events and color videos named HighREV, which provides a challenging evaluation setting for the examined task.  Extensive experiments show that our network consistently outperforms previous state-of-the-art methods on frame interpolation, single image deblurring, and the joint task of both. Experiments on domain transfer reveal that self-supervised training effectively mitigates the performance degradation observed when transitioning from synthetic data to real-world data. Code and datasets are available at \url{https://github.com/AHupuJR/REFID}.}
\end{abstract}

\begin{IEEEkeywords}
Event camera, video frame interpolation, motion deblurring, self-supervised learning, low-level vision
\end{IEEEkeywords}}
\maketitle

\IEEEdisplaynontitleabstractindextext

% \fi
\IEEEpeerreviewmaketitle

\IEEEraisesectionheading{\section{Introduction}\label{sec:intro}}

\IEEEPARstart{V}{ideo} frame interpolation (VFI) methods synthesize intermediate frames between consecutive input frames, increasing the frame rate of the input video, with wide applications in super-slow generation~\cite{jiang2018super,niklaus2020softmax,hu2022many}, video editing~\cite{ren2018deep,zitnick2004high}, virtual reality~\cite{anderson2016jump}, and video compression~\cite{wu2018video}. With the absence of inter-frame information, frame-based methods explicitly or implicitly utilize motion models such as linear motion~\cite{jiang2018super} or quadratic motion~\cite{xu2019quadratic}. However, the non-linearity of motion in real-world videos makes it hard to accurately capture inter-frame motion with these simple models.

\begin{figure}[!tb]
    \centering 
    \includegraphics[width=0.48\textwidth]{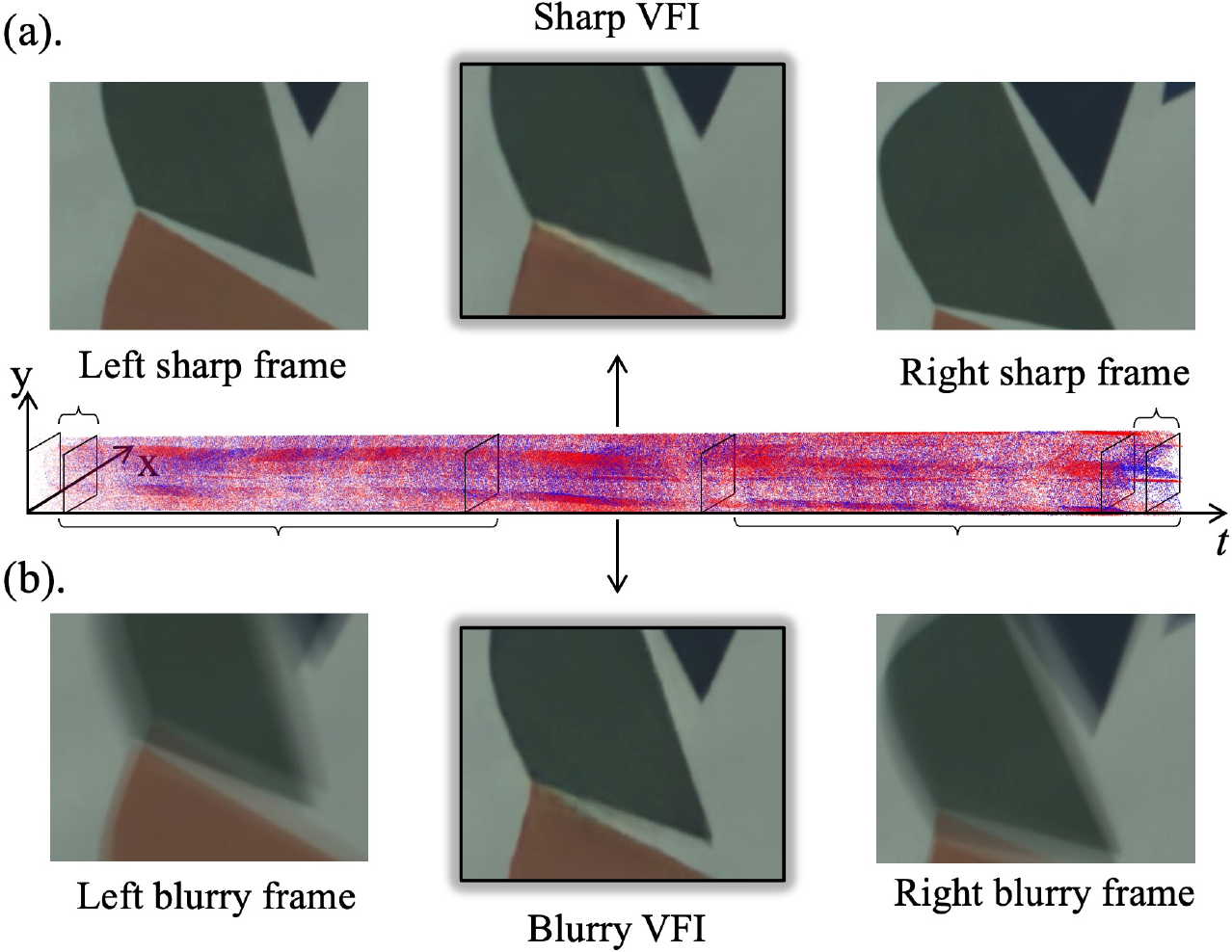}
    \caption{Our unified framework for event-based sharp VFI (a) and blurry VFI (b). Red/blue dots: negative/positive events; Curly braces: exposure time range.}
    \label{fig:tizer}
\end{figure}

%%%% introduce event-based frame interpolation
Recent works introduce event cameras in VFI as a proxy to estimate the inter-frame motion between consecutive frames. Event cameras~\cite{gallego2020event,bao2024temporal,chen2022efficient} are bio-inspired asynchronous sensors that report per-pixel intensity changes, \ie, \textit{events}, instead of synchronous full intensity images. The events are recorded at high temporal resolution (in the order of $\mu$s) and high dynamic range (over 140 dB) within and between frames, providing valid compressed motion information. Previous works~\cite{tulyakov2021time,tulyakov2022time,he2022timereplayer,sun2023event} show the potential of event cameras in VFI, comparing favorably to frame-only methods, especially in high-speed non-linear motion scenarios, by using spatially aligned events and RGB frames.
%%%% illustrate the key problem (motion blur) in event-based high-speed VFI
These event-based VFI methods make the crucial assumption that the input images are sharp. However, this assumption is violated in real-world scenes because of the ubiquitous motion blur. In particular, because of the finite exposure time of frames in real-world videos, especially of those captured with event cameras that output both image frames and an event stream (\ie, Dynamic and Activate VIsion Sensor (DAVIS)~\cite{brandli2014240})---which have a rather long exposure time and low frame rate, motion blur is inevitable for high-speed scenes. In such a scenario, where the reference frames for VFI are degraded by motion blur, the performance of frame interpolation also degrades.

%%%%%% introduce event-based deblur and frame interpolation.
As events encode motion information within and between frames, several studies~\cite{lin2020learning_event_video_deblur,edi_pan,haoyu2020learning} are carried out on event-based deblurring in conjunction with VFI. However, these works approach the problem via cascaded deblurring and interpolation pipelines and the performance of VFI is limited by the image deblurring performance.

%%%% Our goal: the two task can be unified with event camera.
Thus, the desideratum in event-based VFI is robust performance on both sharp image interpolation and blurry image interpolation. Frame-based methods~\cite{shen2020blurry,zuckerman2020across,oh2021demfi,huang2020rife,ifrnet,ifrnet} usually treat these two aspects as separate tasks. Different from frames, events are not subject to motion blur. No matter whether the frame is sharp or blurry, the corresponding events are the same. Based on this observation,
% With event as an additional modality information,
we propose to unify the two aforementioned tasks into one problem: \textit{given two input images and a corresponding event stream, restore the latent sharp images at arbitrary times between the input images.} The input images could be either blurry or sharp, as Fig.~\ref{fig:tizer} shows.
%%%% intro our method
\Add{To solve this problem, we first revisit the physical model of event-based deblurring and frame interpolation. Based on this model, we propose a novel recurrent network, named \textbf{R}ecurrent \textbf{E}vent-based \textbf{F}rame \textbf{I}nterpolation with ad-hoc \textbf{D}eblurring (REFID), which can perform event-based sharp VFI, event-based blurry VFI, and single image deblurring.} The network consists of two branches, an image branch and an event branch. The recurrent structure pertains to the event branch, in order to enable the propagation of information from events across time in both directions. Features from the image branch are fused into the recurrent event branch at multiple levels using a novel attention-based module for event-image fusion, which is based on the squeeze-and-excitation operation~\cite{hu2018squeeze}.

%%% introduce the ssl part
\Add{In the realm of event-based image and video deblurring (\ie \textit{single image deblurring} and \textit{blurry frame interpolation} in our work), a significant challenge lies in achieving robust generalization to real-world conditions~\cite{rebecq2018esim,hu2021v2e}. The distinctive characteristics arising from hardware limitations, sensor noise, and uncertainties in parameters such as the threshold ($c$ in \eqref{eq:event_emit}) contribute to a domain gap between synthetic events and those captured by specific event cameras. While some researchers address this issue by collecting real-world data~\cite{sun2022event}, the impracticality of obtaining ground truth for each event camera poses a substantial limitation.
To tackle this challenge, we propose a novel approach involving the fine-tuning of the model on real-world data, leveraging a self-supervised learning methodology. This approach facilitates generalization across different cameras without the need for collecting ground truth. Within the framework of our proposed self-supervised learning approach, we integrate various constraints throughout the image degradation process and delve into the exploration of motion compensation in the domain of event-based deblurring.}

%%%%% dataset
To test our method on a real-world setting and motivated by the lack of event-based datasets recorded with high-quality event cameras, we record a dataset, \ourdataset, with high-resolution chromatic image sequences and corresponding events. From the sharp image sequences, we synthesize blurry images by averaging several consecutive frames~\cite{nah2017deep}. To our knowledge, \ourdataset has the highest resolution in both image and event among all publicly available event datasets.

In summary, we make the following contributions:
\begin{compactitem}
\item We propose a framework for solving general event-based frame interpolation and event-based single image deblurring,
% event-based frame interpolation, event-based motion deblurring, and both of them jointly, 
which builds on the underlying physical model of high-frame-rate video frame formation and event generation.
\item We introduce a novel network for solving the above tasks, which is based on a bi-directional recurrent architecture, includes an event-guided channel-level attention fusion module that adaptively attends to features from the two input frames according to the temporal proximity with features from the event branch, and achieves state-of-the-art results on both synthetic and real-world datasets.

\item \Add{We integrate REFID with a self-supervised framework with motion compensation for event-based image/video deblurring. This utilizes the event generative model and constraints between the blurry images and a sharp video clip.}

% \item \Add{In this framework, motion is estimated from events, and used to warp the events into sharp edge-maps which improve the result compared with voxel grid representation.}

\item\Add{We evaluate the proposed framework on benchmarks with synthetic events and real events in self-supervised learning settings. The integration of REFID with a self-supervised fine-tuning framework allows for model refinement using real data, even in the absence of ground truth.}

\item We present a new real-world high-resolution dataset with events and RGB videos, which enables real-world evaluation of event-based interpolation and deblurring.

\end{compactitem}

\section{Related Work}
\label{sec:related_work}

\subsection{Event-based frame interpolation} 
Because event cameras report the per-pixel intensity changes, they provide useful spatio-temporal information for frame interpolation. Tulyakov~\etal~\cite{tulyakov2021time} propose Time Lens, which combines a warping-based method and a synthesis-based method with a late-fusion module. Time Lens++~\cite{tulyakov2022time} further improves the efficiency and performance via computing motion splines and multi-scale fusion separately. TimeReplayer~\cite{he2022timereplayer} utilizes a cycle-consistency loss as supervision signal, making a model trained on low-frame-rate videos also able to predict high-speed videos.
All the methods above assume that the key frame is sharp, but in high-speed or low-illumination scenarios, the key frame inevitably gets blurred because of the high-speed motion within the exposure time, where these methods failed (Tab.~\ref{tab:deblurinterpo}). Hence, the exposure time should be taken into consideration in real-world scenes.

\subsection{Event-based deblurring} 
Due to the high temporal resolution, event cameras provide motion information within the exposure time, which is a natural motion cue for image deblurring. Thus, several works have focused on event-based image deblurring.
Jiang~\etal~\cite{jiang2020learning} used convolutional models and mined the motion information and edge information to assist deblurring.
Sun~\etal~\cite{sun2022event} proposed a multi-head attention mechanism for fusing information from both modalities, and designed an event representation specifically for the event-based image deblurring task. 
Kim~\etal~\cite{kim2021event} further extended the task to images with unknown exposure time by activating the events that are most related to the blurry image.
These methods only explore the single image deblurring setting, where the timestamp of the deblurred image is in the middle of the exposure time. However, the events encode motion information for the entire exposure time, and latent sharp images at arbitrary points within the exposure time can be estimated in theory.

\begin{figure*}[!htbp]
    \centering 
    \includegraphics[width=1.\textwidth]{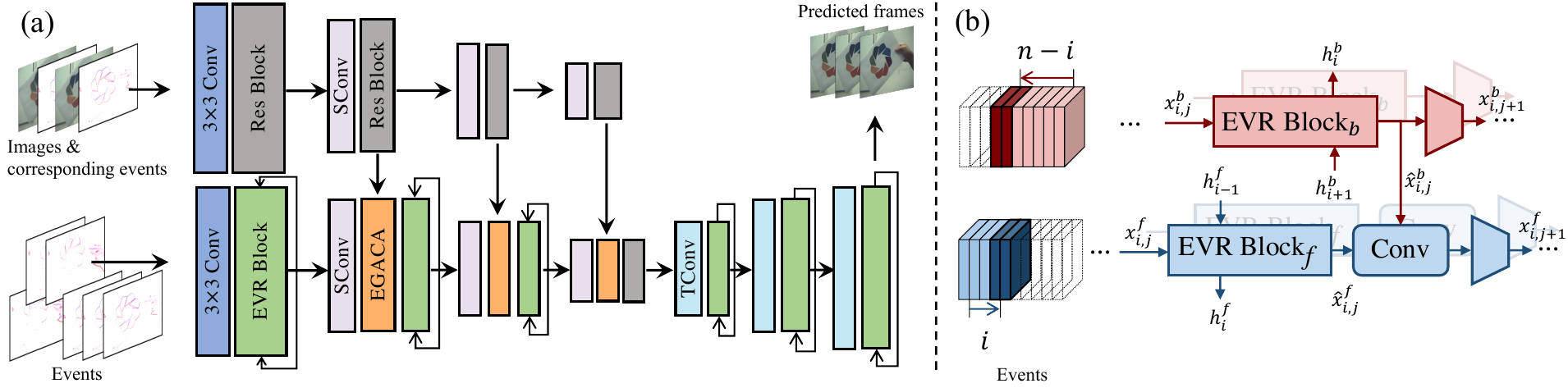}
    \caption{(a): {The architecture of our Recurrent Event-based Frame Interpolation with ad-hoc Deblurring (\ourmodel) network.} The input of the image branch consists of two key frames and their corresponding events, and the event branch consumes sub-voxels of events recurrently. ``EGACA'': event-guided adaptive channel attention, ``SConv'': strided convolution, ``TConv'': transposed convolution. (b): {The proposed bidirectional event recurrent (EVR) blocks.} In each recurrent step, the events from the forward and backward direction are fed to the network. For notations, cf.\ \eqref{eq:evr_block}.}
    \label{fig:model}
\end{figure*}

\subsection{Joint frame interpolation and enhancement} 
Pan~\etal~\cite{edi_pan} formulate the Event Double Integral (EDI) deblurring model, which is derived from the definition of image blur and the measurement mechanism of event cameras, and perform both image deblurring and frame interpolation by accumulating events and applying the intensity changes within the exposure time and from the key frame to the synthesized frames, respectively.
This seminal work optimizes the model by minimizing an energy function but is limited by practical issues in the measurement mechanism of event cameras, e.g.\ accumulated noise, dynamic contrast thresholds and missing events.
Based on EDI, a differentiable model and a residual learning denoising model to improve the result is introduced in~\cite{wang2019event}. Recent works~\cite{zhang2022unifying,paredes2021back,bao2023improving,sun2023event} identify the relationship between the events and the latent sharp image, and apply it to self-supervised event-based image reconstruction and image deblurring.
However, the above works on joint frame interpolation and deblurring predict the latent frames with a two-stage deblurring+interpolation approach, which limits the performance of VFI.

\subsection{\Add{Self-supervised deblurring}}
% Traditionally, deblurring algorithms model the image formation as the convolution of a blur kernel with a sharp image, which is then estimated by means of deconvolution
% The blur kernel, however, is not straightforward to estimate. A common assumption is that the kernel is spaceinvariant [6, 22, 15], which is only valid when the scene is static—camera shake is the only source of motion blur—and planar. Even under this simplifying assumption, the problem is severely ill-posed, and thus requires regularization: , TV [1], heavy-tailed gradient distribution [20], Gaussian distribution [levin2007image], smoothness [shan2008high], and for the shape of the kernel
% pan2014deblurring effective L0-regularized prior based on intensity and gradient for text image deblurring

% optimization methods
\Add{
Traditional deblurring methods model the image degradation process as a convolution of a blur kernel with a latent sharp image, in which estimating accurate blur kernels is essential to the result. 
Other researchers dedicated lots of efforts to designing regularizations to deal with the highly ill-posed problem: total-variance~\cite{bioucas2006total}, Gaussian distribution~\cite{levin2007image}, intensity and gradient prior of text images~\cite{pan2014deblurring}, \etc. However, these assumptions may fail in complex real-world scenarios. Other works~\cite{fergus2006removing,levin2009understanding,shan2008high,chakrabarti2010analyzing,dai2008motion} attempted to remove the dependence on assumptions by utilizing spatially variant blur kernels. However, these methods fail to account for non-planar and object-moving scenes, making them unsuitable for practical use.
Chen~\etal~\cite{chen2018reblur2deblur} predicted the optical flow using latent images from kernel-free estimation and re-render the blurry image, and combining self-supervised losses with ground truth for a hybrid training to improve the deblurring results.
Liu~\etal~\cite{liu2020self} constrained the self-supervised image deblurring with the linear motion assumption in the exposure time and proposed a differentiable model to complete the blur consistency with predicted optical flow and latent sharp images. 
Based on \cite{liu2020self}, Yu~\etal~\cite{yu2023learning} proposed to utilize events to predict the optical flow within the exposure time of the image, and performed single image deblurring. In contrast to Yu et al.~\cite{yu2023learning}, who solely considered events within the exposure time, our methods incorporate an event-generation model and leverage event information both within and beyond the exposure time.}

\section{Our Approach}
\label{sec:method}

We first revisit the physical model of event-based frame interpolation and deblurring in Sec.~\ref{subsec:problem}. Based on this model, we argue that the events within the exposure time should not be ignored in event-based frame interpolation, and present our model architecture abstracted from the physical model in Sec.~\ref{subsec:general_architecture}. To perform the bidirectional recurrent propagation, we demonstrate the data preparation in Sec.~\ref{subsec:data_preparation}. In Sec.~\ref{subsec:evb} and Sec~\ref{subsec:attention}, we introduce the proposed bidirectional Event Recurrent Block and Event-Guided Adaptive Channel Attention in detail. \revise{Note that all the symbols are summarized in Tab.~\red{8} in the Supplementary Materials.}

\subsection{Problem Formulation}
\label{subsec:problem}
% introduce event camera
Once the change in intensity $\mathcal{I}$ at a pixel between the current moment and the moment of the last generated event at that pixel surpasses the contrast threshold $c$, an event camera emits the $i$-th event $e_{i}$, represented as a tuple $e_i = (x_i, y_i, t_i, p_i)$, where $x_i$ and $y_i$ represent the pixel coordinates of the event, $t_i$ represents its timestamp, and $p_i$ is the polarity of the \revise{single} event. More formally, this can be written as
% \begin{equation}
% \label{eq:event_emit}
% p_i = 
% \begin{cases}
% +1, \text{if} \log \left( \frac{\mathcal{I}_t {(x_i, y_i)}} {\mathcal{I}_{ t - \Delta t} {(x_i, y_i)} } \right) > c, \\
% -1, \text{if} \log \left( \frac{\mathcal{I}_t {(x_i, y_i)}} {\mathcal{I}_{ t - \Delta t} {(x_i, y_i)} } \right) < -c.
% \end{cases}
% \end{equation}

\begin{equation}
\label{eq:event_emit}
p_i = 
\begin{cases}
+1, \text{if} \log \left( \frac{\mathcal{I}_t {(x_i, y_i)}} {\mathcal{I}_{ t - \Delta t} {(x_i, y_i)} } \right) > c, \\
-1, \text{if} \log \left( \frac{\mathcal{I}_t {(x_i, y_i)}} {\mathcal{I}_{ t - \Delta t} {(x_i, y_i)} } \right) < -c.
\end{cases}
\end{equation}

Ideally, given two consecutive images, referred to as the left frame $I_{0}$ and the right frame $I_{1}$, and the corresponding event stream in the time range between the timestamps of the two images $[t_{0}, t_{1}]$, we can get any latent image $\hat{I}_{\tau}$ with timestamp $\tau$ in $[t_{0},t_{1}]$ via
\begin{equation}
    \begin{aligned}
    \hat{{I}}_{\tau} = {I}_{0} \operatorname{exp}(c\int_{t_{0}}^t p(s)ds),  \\
    \hat{{I}}_{\tau} = {I}_{1} \operatorname{exp}(c\int_{t_{1}}^t p(s)ds),
    \end{aligned}
    \label{eq:interpo}
\end{equation}
\revise{
where $p(s)$ is the polarity component of the event stream. Note that in \eqref{eq:interpo}, the $p(s)$ is the set of all the polarities of the event stream, with the same dimension as the event image.}

Previous event-based methods~\cite{tulyakov2021time, tulyakov2022time,he2022timereplayer} solve event-based frame interpolation based on \eqref{eq:interpo}.
However, in the real-world setting, because of the finite exposure times of the two frames,
the timestamps $t_{0}$ and $t_{1}$ should be replaced by time ranges, and the images $I_{0}$ and $I_{1}$ may be either sharp (small motion in the exposure time) or blurry (large motion in the exposure time).
% Thus, exposure time should not be ignored in the event-based frame interpolation task, especially in the high-speed scenes, where motion blur usually appears.
Thus, the events within the exposure time of the frames, $e$, should also be utilized for removing potential blur from the frames:

\begin{equation}
    \text{Deblur}(I,e) = \frac{B \times T}{\int_{t_{s}}^{t_{e}} \exp \Big (c \int_{\frac{t_{s}+t_{e}}{2}}^{t} p(s) ds\Big)\ dt,}
    \label{eq:deblur}
\end{equation}
where $B$, $T$, $t_{s}$ and $t_{e}$ are the blurry frame, length of exposure time, start and end of exposure time, respectively. Previous studies~\cite{haoyu2020learning,edi_pan,lin2020learning_event_video_deblur,jiang2020learning} combine the above deblur equation with the frame interpolation equation \eqref{eq:interpo} (denoted as $\text{Interpo}$) to synthesize the target frame:
\begin{equation}
    \begin{aligned}
    \hat{I}_{\tau,0} = \text{Deblur}(I_{0}, \epsilon_{t_{0,s}\rightarrow{}t_{0,e}})\text{Interpo}(\epsilon_{t_{0,s}\rightarrow\tau}),  \\
    \hat{I}_{\tau,1} = \text{Deblur}(I_{1}, \epsilon_{t_{1,s}\rightarrow{}t_{1,e}})\text{Interpo}(\epsilon_{\tau\leftarrow{}t_{1,e}}),
    \end{aligned}
    \label{eq:interpo_with_T}
\end{equation}

% E_{\tau\leftarrow{}t_{1,e}}

where $\epsilon_{t_{0,s}\rightarrow\tau}$ and $\epsilon_{\tau\leftarrow{}t_{1,e}}$ indicate the intensity changes---recorded as events---from the start of the exposure time of the left frame, $t_{0,s}$, and the end of the exposure time of the right frame, $t_{1,e}$, to the target timestamp $\tau$.

However, the physical model \eqref{eq:interpo_with_T} is prone to sensor noise and to the varying contrast threshold of an event camera, which is an inherent drawback of such cameras.

Based on \eqref{eq:interpo_with_T}, \cite{lin2020learning_event_video_deblur,zhang2022unifying,edi_pan,haoyu2020learning} design deep neural networks with a cascaded \textit{first-deblur-then-interpolate} pipeline to perform blurry frame interpolation. In these two-stage methods, the performance of frame interpolation (second stage) is limited by the performance of image deblurring (first stage). Moreover, these methods are only evaluated on blurry frame interpolation.

Given the left and right frame, we design a unified framework to perform event-based frame interpolation both for sharp and blurry inputs with a one-stage model, which applies deblurring ad-hoc.

\subsection{General Architecture}
\label{subsec:general_architecture}

The physical model of \eqref{eq:interpo_with_T} indicates that the latent sharp frame at time $\tau$ can be derived from the two consecutive frames and the corresponding events as
\begin{equation}
    \begin{aligned}
    \hat{I}_{\tau,0} = \mathbf{F}(\mathbf{G}(I_0,E_0),E_{t_{0,s}\to\tau}),  \\
    \hat{I}_{\tau,1} = \mathbf{F}(\mathbf{G}(I_1,E_1),E_{\tau\leftarrow{}t_{1,e}}), \\
    \end{aligned}
    \label{eq:interpo_abstract}
\end{equation}
where $\mathbf{G}$ and $\mathbf{F}$ are learned parametric mappings, \revise{representing ``Deblur'' and ``Interpo'' function in \eqref{eq:interpo_with_T}. Contrary to the formulation of \eqref{eq:interpo_with_T}, $\mathbf{G}$ does not accomplish solely image deblurring, but rather extracts features of both absolute intensities (image) and relative intensity changes (events) within the exposure time. We use cascaded residual blocks to model this mapping, \ie, ``Res Block'' in Fig.~\ref{fig:model}}. For each latent frame, previous methods collect the events in both time ranges and convert them to an event representation~\cite{tulyakov2021time,he2022timereplayer}, which may incur inconsistencies in the result~\cite{tulyakov2022time}. To mitigate this, we use a recurrent network to naturally model temporal information. Thus, we abstract the physical model \eqref{eq:interpo_with_T} to:
\begin{equation}
    \begin{aligned}
    \hat{I}_{\tau,0} = {\textbf{EVR}}_f(\mathbf{G}(I_{0}, E_{0}, I_{1}, E_{1}), E_{\tau}, E_{t_{0,s}\rightarrow\tau}),  \\
    \hat{I}_{\tau,1} = {\textbf{EVR}}_b(\mathbf{G}(I_{0}, E_{0}, I_{1}, E_{1}), E_{\tau}, E_{\tau\leftarrow{}t_{1,e}}),  \\
    \end{aligned}
    \label{eq:recurrent_abstract}
\end{equation}
where ${\textbf{EVR}}_f$ and ${\textbf{EVR}}_b$ denote forward and backward event recurrent (EVR) blocks, respectively. \eqref{eq:recurrent_abstract} summarizes the architecture of our proposed method. $E_{\tau}$ refers to the events in a small time range centered around $\tau$. The recurrent blocks accept as input not only current events, but also previous event information through their hidden states.

\begin{figure}[t]
    \centering
    \includegraphics[width=0.45\textwidth]{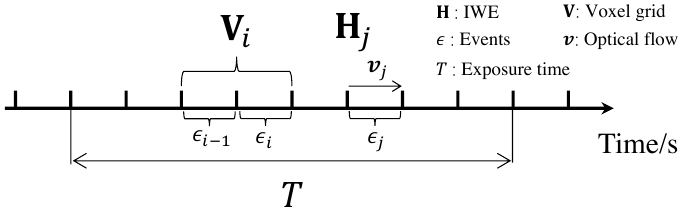}
    \vspace{-5pt}
    \caption{\revise{Details of network inputs. Events within the exposure time $T$ and the blurry frame are unfolded into $N$ sharp images. Events are split into sub-intervals $\epsilon_i$, and two sub-intervals of events are used to compute 2-channel voxel grids $\textbf{{V}}_i$. $\epsilon_i$ is also used to predict optical flow $\textbf{u}_j$. Events are warped to produce IWE $\textbf{H}_j$ with $\textbf{u}_j$ for each sub-interval.}}
    % Images of warped events $\textbf{H}_i$, and flows $\textbf{v}_i$.}}
    \label{fig:network_input}
\end{figure}

Because of the sensor noise and the varying contrast threshold of the event camera sensor, $\hat{I}_{\tau,0}$ ($\hat{I}_{\tau,1}$) approximates the latent sharp image more accurately when the corresponding timestamp of the latter, $\tau$, is closer to $t_{0}$ ($t_{1}$).
To fuse $\hat{I}_{\tau,0}$ and $\hat{I}_{\tau,1}$ implicitly, we further propose a new Event-Guided Adaptive Channel Attention (\ourfusionmodule) module to mine and fuse the features from the image branch of REFID with adaptive weights determined by the current events:

\begin{equation}
    \hat{I}_{\tau} = \text{Fuse}(\hat{I}_{\tau,0}, \hat{I}_{\tau,1}).
    \label{eq:fusion_abstract}
\end{equation}

The overall network architecture of REFID is shown in Fig.~\ref{fig:model} (a). The image branch extracts features from the two input images and the corresponding events and is connected to the event branch at multiple scales. Overall, \ourmodel has a U-Net~\cite{ronneberger2015u} structure. A bidirectional recurrent encoder with EVR blocks extracts features from current events and models temporal relationships with previous and future events. In each block of the encoder, the features from the image branch are fused with those from the event branch adaptively with our novel \ourfusionmodule module, which we detail in \ref{subsec:attention}.

\Add{The proposed REFID can be extended to single-image deblurring by utilizing a single frame and its corresponding events for the image and event branches, respectively. Moreover, our approach generates multiple latent sharp images as opposed to only one, as highlighted in the work by Sun et al.~\cite{sun2022event}.}

\begin{figure*}[t]
    \centering 
    \includegraphics[width=1.\textwidth]{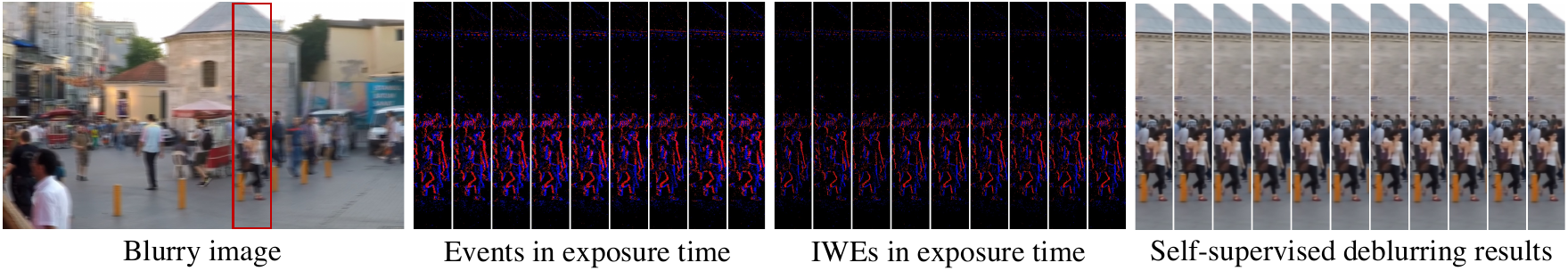}
    % \vspace{-6mm}
    \caption{A example for self-supervised single-image deblurring, a sharp video clip is restored with a blurry image and corresponding events. From left to right: Visualized events, image of warped events (IWE), and resulting sharp video clip. IWE provides sharper edge information while events contribute to capturing the blurry shape information.}
    \label{fig:iwe}
    % \vspace{-4mm}
\end{figure*}

\subsection{Data Preparation}
\label{subsec:data_preparation}
To feed the asynchronous events to our network, we first need to convert them to a proper representation. \Add{The detailed model inputs are depicted in Fig.~\ref{fig:network_input}}. According to \eqref{eq:recurrent_abstract}, the latent image can be derived in both temporal directions. Thus, apart from the forward event stream, we reverse the event stream both in time and polarity to get a backward event stream. 
Then, event streams from the two directions are converted to two voxel grids~\cite{Rebecq19cvpr,rebecq2019events}, \revise{and we take one voxel grid $\textbf{{V}}_{total}\in \mathbb{R}^{(n+2)\times H \times W}$ for an example, where $n$ is the number of interpolated frames}
{\Add{$V_{i}\in \mathbb{R}^{(n+2)\times H \times W}$} and 
$V\in \mathbb{R}^{(n+2)\times H \times W}$, where $n$ is the number of interpolated frames.} The channel dimension of the voxel grids holds discrete temporal information. \Add{In each recurrent iteration, $\textbf{{V}}_{\text{i}}\in \mathbb{R}^{2\times H \times W}$ are constructed from small sub-intervals of events $\epsilon_i = \{e_k \vert \tau_i\leq t_k \leq\tau_{i+1}\}$. \revise{$\textbf{{V}}_{\text{i}}$ from both directions} are fed to the event branch, which encodes the event information for the latent frame.} We also convert events in the exposure time of the two images to voxel grids and concatenate them with corresponding images to form the input of the image branch. 

% \sout{Further, for self-supervised fine-tuning experiments, each group of events is also used to compute an optical flow segment in self-supervised settings $\mathbf{v}_i(\textbf{x})$ following the model-based method~\cite{shiba2022secrets} for event-based optical flow estimation and a 2-channel image of warped events (IWE) $\mathbf{H}_i(\textbf{x})$ with the estimated optical flow. For IWE we compute the projection of motion-compensated events via}

\revise{Further, for self-supervised fine-tuning experiments, each group of events is also used to compute an optical flow segment 
in self-supervised settings $\mathbf{v}(\textbf{x}_k)$ following the model-based method~\cite{shiba2022secrets} for event-based optical flow estimation and a 2-channel image of warped events (IWE) $\mathbf{H}_i(\textbf{x})$ with the estimated optical flow, where $\textbf{x}\doteq(x,y)^\top$. For IWE we compute the projection of motion-compensated events via}
\begin{align}
    \mathbf{H}_i(\mathbf{x}) &= \sum_{e_k\in \epsilon_i} \delta(p-p_k)\delta(\mathbf{x} - \mathbf{u}'_k), \\
    \label{eq:iwe}
    \mathbf{u}'_k &= \mathbf{u}_k - \mathbf{v}_i(\mathbf{u}_k)(t_k - \tau_i),
\end{align}
% \begin{equation}
%     \mathbf{x}'_k = \mathbf{x}_k + (t_k - \tau_i)\mathbf{u}(\mathbf{x}_{k}),
%     \label{eq:iwe}
% \end{equation}
\revise{where $\delta$ represents the Kronecker delta.}
\Add{With the equation we temporally align all events with the interval to the timestamp $\tau_i$\footnote{Note, that for the last interval $\epsilon_{N-1}$, we generate two images of warped events $\mathbf{H}_{N-1}(\mathbf{x})$ and $\mathbf{H}_N(\mathbf{x})$, by once backward warping, and then forward warping the same events.}. Note that the result is a 2-channel tensor, where events are separated according to their polarity. As Fig.~\ref{fig:iwe} shows, the image of warped events provides a strong inductive bias for our network by showing it where sharp edges are to be expected.  To compute the 2-channel voxel grid input $\textbf{{V}}_i(\textbf{x})$, we concatenate two 1-channel voxel grids computed from events $\epsilon_{i-1}$ and $\epsilon_{i}$\footnote{Note that this includes two channels that go beyond the exposure interval, $\epsilon_0$ and  $\epsilon_N$}. Before passing to the network, we concatenate the voxel grid and image of warped events, resulting in a 4-channel input, and apply input normalization.}

\subsection{Bidirectional Event Recurrent Block}
\label{subsec:evb}
%% drawback of previous method
In previous event-based works~\cite{tulyakov2021time,tulyakov2022time,he2022timereplayer}, for each latent sharp image, the events from both left and right images to the target image are accumulated and converted to an event representation. However, compared to the temporal resolution of events, the length of the exposure time of frames is large and not negligible, so simple accumulation from a single timestamp in the above works loses information and is not reasonable. Moreover, inference for different latent frames is segregated, which leads to inconsistencies in the results~\cite{tulyakov2022time}.
%%% our method 
To deal with these problems, we propose a recurrent architecture that models the temporal information both within the exposure time of each frame and between exposure times of different frames. By adopting recurrent blocks, frame interpolation is independent from the exposure time of key frames and it can also be performed inside the exposure time. Features propagated through hidden states of the network also guarantee consistency across the predicted frames. Based on \eqref{eq:interpo_abstract}, we design a bidirectional Event Recurrent (EVR) block to iteratively extract features from the event branch.
%%% details
As Fig.~\ref{fig:model} (b) shows, for each direction, the input sub-voxel ($i-1,i$) only consists of two voxels of the input voxel. In the next recurrent iteration, the selected sub-voxel moves forward to the next time ($i,i+1$). For a given recurrent iteration $i$, the forward EVR block cycles for $i$ times and the backward EVR block cycles for $n-i$ times, where $n$ is the index of the latent sharp image at hand:
\begin{equation}
    \begin{aligned}
    % a = b
    & \hat{x}_{i,j+1}^{b}, h_{i}^{b} = \textbf{EVR}_{b}(x_{i,j}^{b}, h_{i+1}^{b}), \\
    & \hat{x}_{i,j+1}^{f}, h_{i}^{f} = \textbf{EVR}_{f}(x_{i,j}^{f}, h_{i-1}^{f}), \\
    & x_{i,j+1}^{b} = \text{Down}(\hat{x}_{i,j+1}^{b}), \\
    & x_{i,j+1}^{f} = \text{Down}(\text{Conv}(\text{Concat}(\hat{x}_{i,j+1}^{b}, \hat{x}_{i,j+1}^{f}))),
    \end{aligned}
    \label{eq:evr_block}
\end{equation}
where $i$ and $j$ are the indices of sub-voxel and scale, respectively. $x$, $h$, $f$ and $b$ denote feature flow, hidden state, forward and backward, respectively.
We select ResNet as the architecture for the EVR block instead of ConvLSTM~\cite{shi2015convolutional} or ConvGRU~\cite{shi2017deep}, because the time range of events between consecutive frames is rather short (cf. Tab~\ref{tab:ablation_architecture}).
In each EVR block, the features from the two directions are fused through convolution and downsampled to half of the original size. The bidirectional EVR blocks introduce the information flow from both directions, which models $E_{t_{0,s}\rightarrow\tau}$ and $E_{\tau\leftarrow{}t_{1,e}}$ in \eqref{eq:interpo_abstract}, helping reduce artifacts by using the information from the end of the time interval (cf.\ Fig.~\ref{fig:blur_interpo}).

\begin{figure}[!t]
    \centering 
    \includegraphics[width=0.48\textwidth]{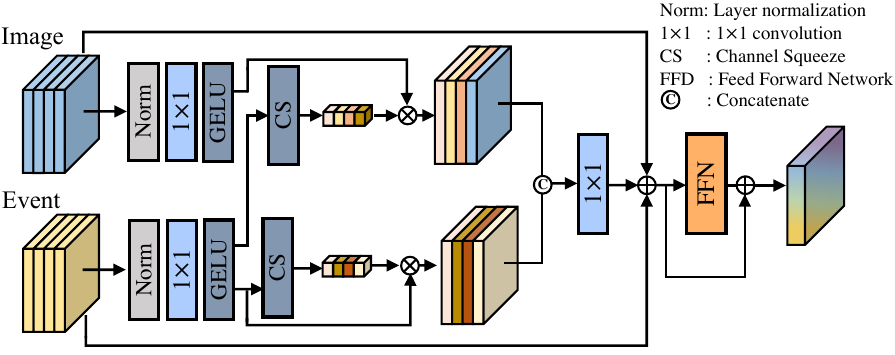}
    \caption{{The Event-Guided Adaptive Channel Attention module.} The channel weights for the image branch are extracted from the event branch.}
    \label{fig:atten}
\end{figure}

\subsection{Event-Guided Adaptive Channel Attention}
\label{subsec:attention}
%%% the fusion is important
In event-based frame interpolation, fusion happens both between the two input frames and between frames and events. Because of the inherent noise of event cameras, the longer the time range between the key frames is, the more the noise in the event accumulation increases. Ideally, the key frame that is closer to the latent frame should contribute more to the prediction of the latter. In other words, the weights of two key frames should be decided by \textit{time}. 

%%% the weight of two frames should decided by the current event sub-voxel.
In our \ourmodel network, the two key frames and the corresponding events are concatenated along the channel dimension to provide the input of the image branch. We design the novel Event-Guided Adaptive Channel Attention (EGACA) module to fuse the two key frames and events at the current input sub-voxel in the recurrent structure. The current input sub-voxel contains events in a small range around the timestamp of the latent frame and the fusion weights for the two key frames and the events are determined by the current input sub-voxel, which indicates the time.

%%% detail
Fig.~\ref{fig:atten} shows the detailed architecture of the proposed EGACA. We simplify the multi-head channel attention of EFNet~\cite{sun2022event} to channel attention from SENet~\cite{hu2018squeeze}. Two Channel Squeeze (CS) blocks extract channel weights from the current events, and two weights multiply event features and image features for self-attention and event-guided attention to image features, respectively. Then, feature maps from the two branches are fused by a feed-forward network. In each recurrent iteration, the channel weights from the current events are different, which helps to mine different features from the two images along the channel dimension.

\subsection{\Add{Self-supervised Framework}}
\label{subsec:losses}

\begin{figure}
    \centering
    %% TODO, change to the first line if final!
    \includegraphics[width=0.45\textwidth]{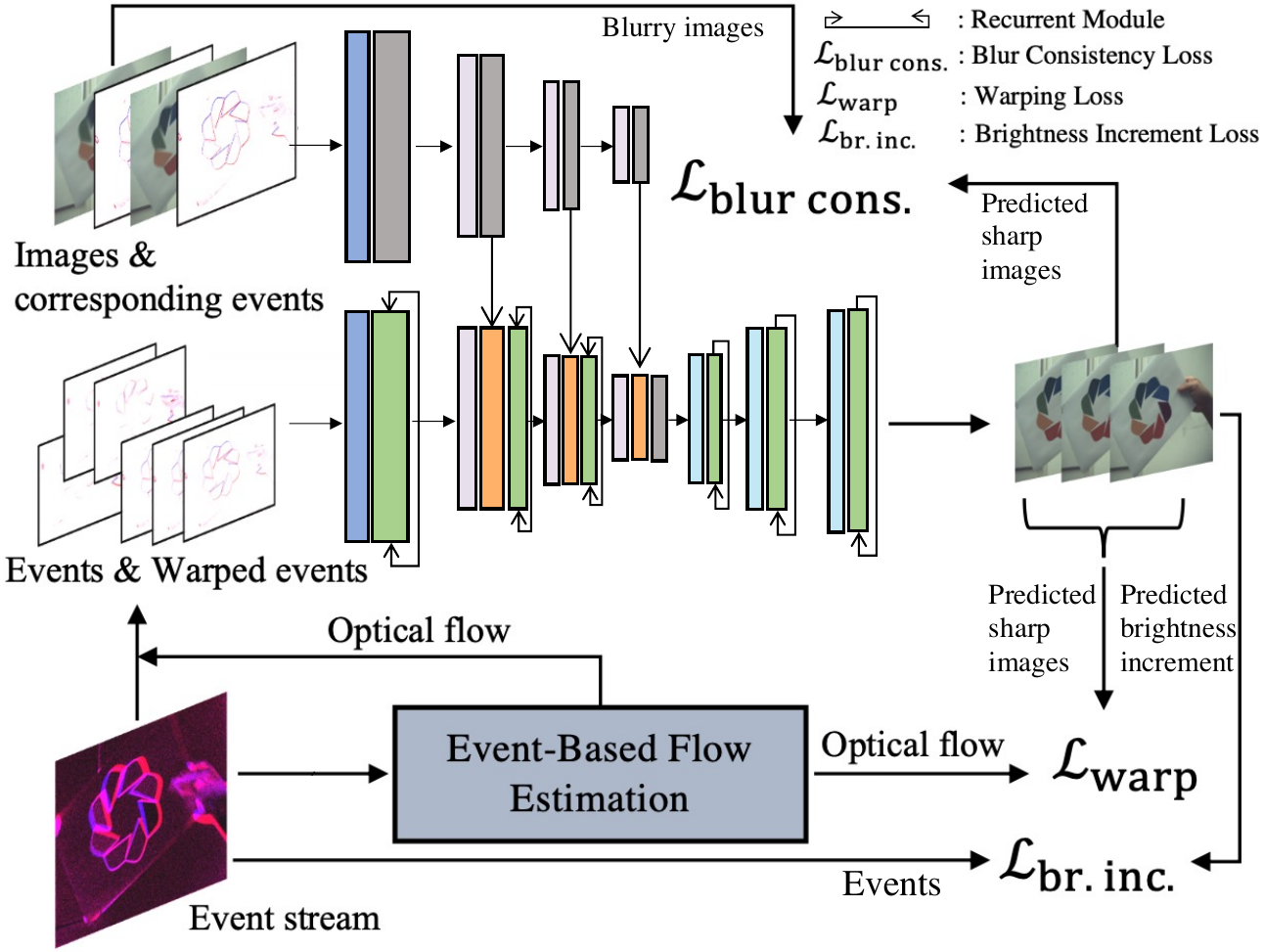}

    \caption{\revise{Self-supervised framework with the same basic architecture illustrated in Fig~\ref{fig:model}: The events are directed to the optical flow estimation module, with the resulting flow utilized for warping events for IWE and computing the Warping Loss with \eqref{eq:warping_loss}. Input blurry images and predicted sharp images are utilized in the calculation of Blur Consistency Loss in \eqref{eq:loss_blur_consistency_final}. Input events and predicted brightness increment are used in the calculation of Brightness Increment Loss in \eqref{eq:loss_brighness_increment_norm}.}}
    \label{fig:ssl_framework}
\end{figure}

\Add{Fig.~\ref{fig:ssl_framework} shows the framework of the self-supervised fine-tuning strategy. In the self-supervised fine-tuning, we find that three self-supervised loss functions are sufficient to train our model, namely (\textit{i}) a brightness increment loss, (\textit{ii}) a blur consistency loss, and (\textit{iii}) a warping loss. While the blur consistency loss is calculated over the entire interval, the other two losses can be defined for each time interval at $\tau_i$ and thus we obtain dense supervision throughout the interval. The resulting loss contains the following $1+2(N-1)$ terms
\begin{equation}
    \label{eq:total_loss}
    \mathcal{L}_\text{tot} = \mathcal{L}_{\text{blur cons.}}+\sum_{i=1}^{N-1}\lambda_0\mathcal{L}^i_{\text{br. inc.}} + \lambda_1\mathcal{L}^i_{\text{warp}},
\end{equation}
and we will now discuss these in turn. Note that all the self-supervised experiments are conducted with single-image deblurring settings
}

\Add{
\textbf{Brightness Increment Loss}
As alluded to in ~\eqref{eq:event_emit}, event cameras approximately measure (up to quantization) the brightness increment between two time instances. We enforce this constraint by minimizing the difference between the measured and predicted log brightness increment between adjacent predicted frames: $\Delta \hat{{L}}_i(\textbf{x})\doteq \log \hat{{I}}_{i+1}({x}) - \log\hat{{I}}_i({x})$.
\begin{equation}
    \label{eq:loss_brighness_increment}
    \mathcal{L}_{\text{br. inc.}} = \Vert \Delta {L}(\textbf{x};\epsilon_i) - \Delta \hat{{L}}_i(\textbf{x}) \Vert_2^2.
\end{equation}
We compute the brightness increment as in \eqref{eq:event_emit}. Since $\Delta {L}$ depends on an unknown contrast threshold $c$, we slightly modify the above loss, to minimize the difference of normalized terms:
\begin{equation}
    \label{eq:loss_brighness_increment_norm}
    \mathcal{L}^i_{\text{br. inc.}} = \left| \frac{\Delta {L}(\textbf{x};\epsilon_i)}{\left\Vert \Delta {L}(\textbf{x};\epsilon_i)\right\Vert_2} - \frac{\Delta \hat{{L}}_i(\textbf{x})}{\Vert \Delta \hat{{L}}_i(\textbf{x})\Vert_2}\right|.
\end{equation}
This modification results in a cancellation of the constant term $c$. Note that this is a common trick employed in works like \cite{gehrig2020eklt,bryner2019event,paredes2021back}. However, different from these works, we do not minimize the first-order linear approximation to the event generation model, but instead use the true model without linearization, and thus are free from approximation error.
}

% Following the approximation in \eqref{eq:motion_blur}

\textbf{Blur Consistency Loss: }
\revise{Theoretically, the intensity of the blurry image equals the average intensity of all the latent sharp images within the exposure time of the blurry image:}
\begin{equation}
    \revise{B(\textbf{x})} \revise{=} \revise{\frac{1}{N}\sum_{i=1}^N {I}_i(\textbf{x}).}
    \label{eq:blur_defination}
\end{equation}
\revise{Based on the image blurring process above, we devise a self-supervised loss that relates the measured blurry frame and predicted sharp frames via }
\begin{equation}
    \mathcal{L}_{\text{blur cons.}} = \left\Vert {B}(\textbf{x}) - \frac{1}{N}\sum_{i=1}^N \hat{{I}}_i(\textbf{x})\right\Vert_2^2.
    \label{eq:loss_blur_consistency}
\end{equation}
\revise{Here all the $I$ are within the exposure time of the blurry image $B$.}
We found that $N=11$ does not generate sufficiently many sharp images to generate realistic blur, and for this reason, \revise{we reuse the optical flow derived from ~\cite{shiba2022secrets} (a non-learning method) directly} to upsample the frames by a factor of $M=4$, generating 3 additional intermediate frames between consecutive sharp images (Note that the event-based flow estimation module is not the contribution of our work, and it can any other event-based optical flow estimation method):
\begin{equation}
    \label{eq:intermediate_frames}
    \hat{{I}}_{i,m}(\textbf{x}) = \hat{{I}}_{i+1}\left(\textbf{x}-\frac{m}{M}{v}_i(\textbf{x})\right), \quad \text{ for } m=1,...,M-1.
\end{equation}
For each frame at time $\tau_i$ we get intensities from the next frame at $\tau_{i+1}$ by rescaling the flow appropriately and applying bilinear sampling.
as a result, the blur consistency loss becomes 
\begin{equation}
    \mathcal{L}_{\text{blur cons.}} = \left\Vert {B}(\textbf{x}) - \frac{1}{NM}\sum_{i=1}^N \sum_{m=0}^{M-1}\hat{{I}}_{i,m}(\textbf{x})\right\Vert_2^2,
    \label{eq:loss_blur_consistency_final}
\end{equation}
where we make use of the fact that $\hat{{I}}_{i,0} = \hat{{I}}_{i+1}$.

\Add{
\textbf{Warping Loss: } %Finally, the above two losses do not capture well regions that have few events, and training only on those resulted in regions that were not deblurred due to a lack of events. However, since we recover dense optical flow $\textbf{v}_i(\textbf{x})$ we may circumvent this issue: 
With the dense optical flow $\textbf{v}_i(\textbf{x})$ recovered previously, we warp the latent sharp image back to the last timestamp and compare the warped image and predicted image. With the dense optical flow, this loss is able to also constrain slight intensity changes.
\begin{equation} \label{eq:warping_loss}
\mathcal{L}^i_\text{warp} = \left\Vert \hat{{I}}_i(\textbf{x}) - \hat{{I}}_{i+1}(\textbf{x}-\mathbf{v}_i(\textbf{x}))\right\Vert_2^2.
\end{equation}
}

\Add{Note that for the blurry frame interpolation task, the 
blur consistency loss is applied to the target frame with the timestamp in the exposure time of either blurry frame. For the single image deblurring, all three losses are applied.}

\section{HighREV Dataset}
\label{sec:dataset}

%% why the dataset?
For event-based low-level tasks, such as event-based image deblurring and event-based frame interpolation, most works evaluate their models on datasets originally designed for image-only methods and having only synthetic events. This is because (1) event cameras are not easy to acquire yet, (2) most event cameras are of low resolution and monochrome~\cite{jiang2020learning,sun2022event,kim2021event}, and (3) high-resolution chromatic datasets~\cite{tulyakov2021time,vitoria2022event} are not publicly available. To fill this gap, we record a high-quality chromatic event-image dataset for training, fine-tuning and evaluating event-based methods for frame interpolation and deblurring.

%% aligned methods
\Add{
In Time Lens\cite{tulyakov2021time}, to construct an event-based high-resolution dataset, the authors combine a synchronized, high-resolution ($1280 \times 720$) event camera with an RGB camera to make a hybrid sensor. However, the alignment of the two sensors introduces error both in the temporal axis and the spatial axis. Our HighREV dataset is collected using one sensor that outputs both events and RGB frames at the same time, with a resolution of $1632 \times 1224$. Because it is a Dynamic and Active VIsion Sensor (DAVIS)~\cite{brandli2014240}, events and RGB images are aligned by design.}

\begin{figure}[!tbp]
    \centering 
    \includegraphics[width=0.48\textwidth]{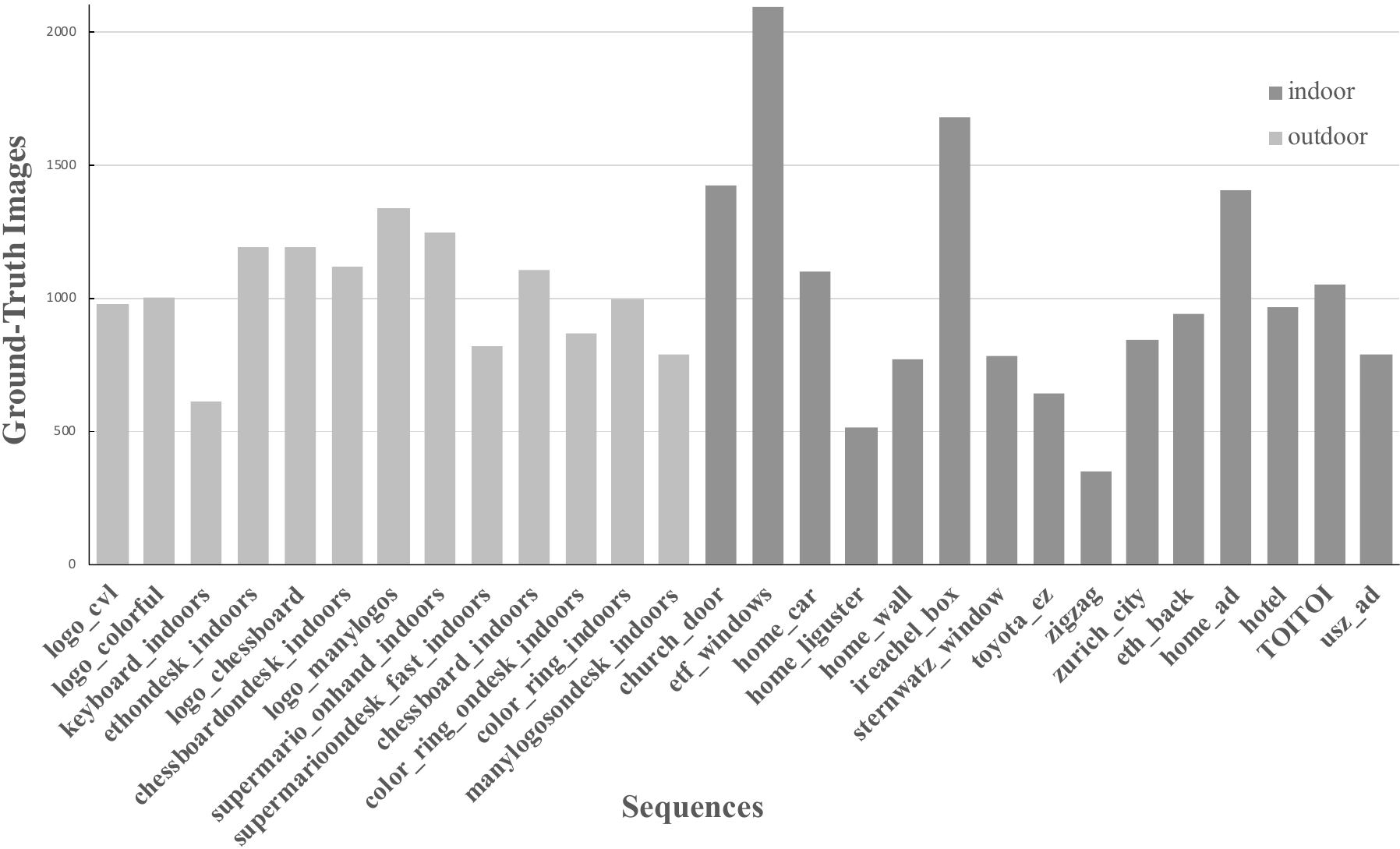}
    \caption{{The distribution of number of ground-truth images per sequence of HighREV dataset.} The x-axis denotes the sequences and y-axis denotes the number of images.}
    \label{fig:dataset}
\end{figure}

%% introduce details
As Fig.~\ref{fig:dataset} shows, our HighREV dataset consists of 30 sequences with a total of 28589 sharp images and corresponding events. We use 19934 images for training/fine-tuning and 8655 images for evaluation. The size of each RGB image is $1632\times1224$. The events and images are spatially aligned in the sensor. Each event has only one channel (intensity), with pixel coordinates, timestamp and polarity. \Add{70\% of the video sequences are used for training and 30\% for testing and we keep the ratio of indoor and outdoor scenes approximately the same in each subset. For the collection of the dataset, the exposure time of the camera is set to 15ms and the f-stop of the lens is set to 2. The frame rate of the APS image is set to 25.}

%% how to make blur
The HighREV dataset can be used for event-based sharp frame interpolation. To evaluate event-based blurry frame interpolation, we synthesize blurry images by averaging 11 consecutive original sharp frames. For blurry frame interpolation, we skip 1 or 3 sharp frames (denoted as \textit{11+1} or \textit{11+3} in Tab.~\ref{tab:deblurinterpo}). To the best of our knowledge, among all event-image datasets, our dataset has the highest resolution.

% \begin{figure*}[!tbp]
%     \centering 
%     % \includegraphics[width=0.48\textwidth]{./figures/dataset.pdf}
%     \includegraphics[width=0.98\textwidth]{./figures/highrev_dataset.pdf}
%     \caption{\textbf{The distribution of number of ground-truth images per sequence of HighREV dataset.} The x-axis denotes the sequences.}
%     \label{fig:dataset}
% \end{figure*}

\section{Experiments}
\label{sec:experiments}

\subsection{Tasks and Datasets}
\label{subsec:datasets}
\Add{For synthetic dataset,} we use the popular GoPro dataset~\cite{nah2017deep} for training and evaluation. GoPro provides blurry images, paired sharp images, and sharp image sequences used to synthesize blurry frames. The images have a size of $1280\times720$. We leverage the event camera simulator ESIM~\cite{rebecq2018esim} to generate simulated event data with threshold $c$ following a Gaussian distribution $\mathit{N}(\mu=0.2, \sigma=0.03)$. \Add{For the real-world dataset, the proposed HighREV dataset is employed.}
For different tasks, the datasets are as follows:

\begin{table*}[th]
\centering
\small
\setlength{\tabcolsep}{10pt}
\caption{{Comparison of sharp frame interpolation methods on GoPro and HighREV.}  ``Frames'' and ``Events'' indicate if a method uses frames and events for interpolation. ``11+1'' (resp.\ ``11+3'') indicates that the blurry image is synthesized with 11 sharp frames and 1 (resp.\ 3) frame(s) is skipped for frame interpolation. The number of network parameters (\#Param) is also provided. \Add{\textit{Ssl.} denotes that the model is self-supervised trained from scratch without ground-truth as supervision.}}
\begin{tabular}{lcccccccc}
\bottomrule[0.15em]
\rowcolor{tableHeadGray}
\textbf{Method} & \textbf{Frames} & \textbf{Events} & \textbf{PSNR $\uparrow$}  & \textbf{SSIM $\uparrow$} & & \textbf{PSNR $\uparrow$} & \textbf{SSIM $\uparrow$}  & \textbf{\#Param} \\ \hline
 \hline
\textbf{GoPro (interpolation)}~\cite{nah2017deep}  &   &    & \multicolumn{2}{c}{\textbf{7 frames skip}}  &     & \multicolumn{2}{c}{\textbf{15 frames skip}}  &    \\ \hline
DAIN~\cite{bao2019depth}                    & \cmark & \xmark & 28.81 & 0.876 &   & 24.39 & 0.736 & 24.0M  \\
SuperSloMo~\cite{jiang2018super}            & \cmark & \xmark & 28.98 & 0.875 &   & 24.38 & 0.747 & 19.8M  \\
IFRNet~\cite{ifrnet}                        & \cmark & \xmark & 29.84 & 0.920 &   & -     & -     & 19.7M  \\
EDI~\cite{edi_pan}                          & \cmark & \cmark & 18.79 & 0.670 &   & 17.45 & 0.603 & 0.5M   \\
TimeReplayer~\cite{he2022timereplayer}      & \cmark & \cmark & 34.02 & 0.960 &   & -  & -  & -     \\

Time Lens~\cite{tulyakov2021time}           & \cmark & \cmark & 34.81 & 0.959 &   & 33.21 & 0.942 & -     \\
\textbf{REFID}                              & \cmark & \cmark & \textbf{36.80} & \textbf{0.980} &  & \textbf{35.635} & \textbf{0.974} & {15.9M} \\ \hline

\textbf{HighREV (interpolation)}  &   &   & \multicolumn{2}{c}{\textbf{7 frames skip}}  &  & \multicolumn{2}{c}{\textbf{15 frames skip}} &   \\ \hline
EDI~\cite{edi_pan}                         & \cmark & \cmark & 22.32 & 0.716 &  & 18.65 & 0.654 & 0.5M \\
RIFE~\cite{huang2020rife}                  & \cmark & \xmark & 32.28 & 0.904 &  & 28.22 & 0.864 & 9.8M  \\
Time Lens~\cite{tulyakov2021time}           & \cmark & \cmark & 32.81 & 0.901 &  & 27.06 & 0.810 & -     \\
\textbf{REFID}                               & \cmark & \cmark & \textbf{38.38} & \textbf{0.977} &  & \textbf{37.58} & \textbf{0.975} & 15.9M  \\ \hline

\end{tabular}
\label{tab:sharpinterpo}
\end{table*}

% \begin{table}[!t]
% \caption{An Example of a Table\label{tab:table1}}
% \centering
% \begin{tabular}{|c||c|}
% \hline
% One & Two\\
% \hline
% Three & Four\\
% \hline
% \end{tabular}
% \end{table}

\begin{figure*}[!tbp]
    \centering 
    \includegraphics[width=0.95\textwidth]{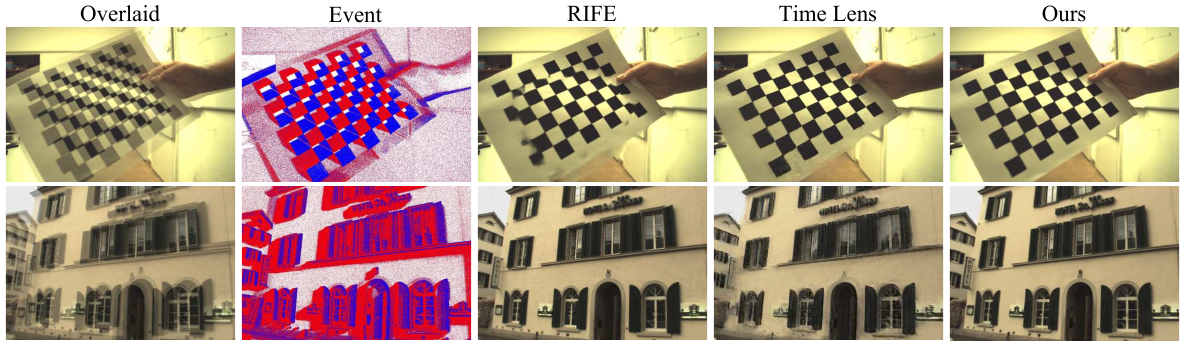}
    \caption{\textbf{Qualitative results for sharp frame interpolation on HighREV.} RIFE~\cite{huang2020rife} suffers from motion ambiguity because of the lack of event information. Time Lens~\cite{tulyakov2021time} is vulnerable to noise. Our \ourmodel shows superior performance both on indoor and outdoor scenes.}
    \label{fig:sharp_interpo}
\end{figure*}

\PAR{Sharp frame interpolation.} The high-frame-rate sharp images of GoPro and HighREV are leveraged by skipping 7 or 15 frames and keeping the next one. \Add{The quantitative results are calculated over all the skipped frames.}

\PAR{Blurry frame interpolation.} We synthesize blurry frames by averaging $11$ sharp high-FPS frames in GoPro and HighREV. Between each blurry frame, we skip 1 or 3 frames for the evaluation of blurry frame interpolation (denoted as ``11+1'' or ``11+3'' in Tab.~\ref{tab:deblurinterpo}). \Add{The metrics of PSNR and SSIM are average quantitative results over all the skipped frames.}

\PAR{Single image deblurring.} We use GoPro with synthesized blurry images (averaged from 7 or 11 sharp frames). For a real-world test, we also fine-tune and evaluate methods on REBlur~\cite{sun2022event}. We only use a single image and its corresponding events in the event branch as input, for a fair comparison.

\PAR{Self-supervised training/fine-tuning.} \Add{We use GoPro and HighREV for self-supervised training or fine-tuning. Experiments on both blurry frame interpolation and single-image motion deblurring are conducted. All the experiments are conducted by self-supervised training except the self-supervised fine-tuning experiment with pre-trained weights (denoted as \textit{pre-trained ssl.} in Tab.~\ref{table:ssl}). Note that the model in self-supervised learning experiments adopts fewer parameters than the model used in supervised setting.}

For blurry frame interpolation and sharp frame interpolation, we train all the models on each training set and evaluate on the respective test set.

\subsection{Implementation Details}
\label{subsec:details}
Different from warping-based methods~\cite{tulyakov2021time, tulyakov2022time}, \ourmodel is an end-to-end network. All its components are optimized from scratch in a single training round, without any pre-trained modules, which makes it train easier. 
We crop the input images and event voxels to $256\times256$ for training and use horizontal and vertical flips, random noise and hot pixels in event voxels~\cite{stoffregen2020reducing}. Adam~\cite{kingma2014adam} with an initial learning rate of $2\times10^{-4}$ and a cosine learning rate annealing strategy with $2\times10^{-4}$ as minimum learning rate are adopted for optimization. We train the model on GoPro with a batch size of 1 for 200k iterations on 4 NVIDIA Titan RTX GPUs. For experiments on HighREV, we fine-tune the model trained on GoPro with an initial learning rate of $1\times10^{-4}$ for 10k iterations. For image deblurring on REBlur, fine-tuning takes $600$ iterations with an initial learning rate of $2\times10^{-5}$.

\Add{
For self-supervised experiments, We consider two settings during training: 
(\textit{i}) self-supervised training from scratch, where we use an initial learning rate of $1\times10^{-4}$, and train for $20,000$ iterations, and 
(\textit{ii}) self-supervised fine-tuning with pre-trained weights, where we use an initial learning rate of $2\times10^{-5}$ and train for $20,000$ iterations. In all settings, we use a final learning rate of $10^{-7}$, and a batch size of 2. All the experiments are conducted on a single Titan RTX GPU.
}

\begin{table*}[th]
\centering
\small
\setlength{\tabcolsep}{12pt}
\caption{{Comparison of blurry frame interpolation methods on GoPro and HighREV.} Read as Tab.~\ref{tab:sharpinterpo}.}
\resizebox{0.95\textwidth}{!}{%
    \begin{tabular}{lcccccccc}
    \bottomrule[0.15em]
    \rowcolor{tableHeadGray}
    \textbf{Method} & \textbf{Frames} & \textbf{Events} & \textbf{PSNR $\uparrow$}  & \textbf{SSIM $\uparrow$} & & \textbf{PSNR $\uparrow$} & \textbf{SSIM $\uparrow$}  & \textbf{\#Param} \\ \hline
     \hline
    \textbf{GoPro}~\cite{nah2017deep}  &   &  &  \multicolumn{2}{c}{\textbf{11+1}}  &       & \multicolumn{2}{c}{\textbf{11+3}}  &                \\ \hline
    \Add{\textbf{REFID (\textit{ssl.})}}  & \cmark & \cmark  & \Add{\textbf{28.61}}  &  \Add{\textbf{0.849}}  &  & \Add{\textbf{26.72}} & \Add{\textbf{0.791}} &  \Add{\textbf{7.8M}}  \\
    RIFE~\cite{huang2020rife}         & \cmark & \xmark  & 28.69 & 0.856 &  & 26.91  & 0.798 & 9.8M \\
    EDI~\cite{edi_pan}                & \cmark & \cmark  & 18.72 & 0.506 &  & 18.49  & 0.486 & 0.5M \\
    Time Lens~\cite{tulyakov2021time}  & \cmark & \cmark  & 21.56 & 0.581 &  & 21.47  & 0.587 & 72.9M  \\
    EVDI~\cite{zhang2022unifying}     & \cmark & \cmark  & 29.17 & 0.880 &  & 28.77  & 0.873 & 0.4M    \\

    EFNet+IFRNet~\cite{sun2022event,ifrnet} & \cmark & \cmark  & 33.05 & 0.955 &  & 32.89  & 0.950 & 28.2M   \\
    E2VID+~\cite{rebecq2019events}    & \cmark & \cmark  & 33.82 & 0.961 &  & 33.39  & 0.954 & 15.3M  \\
    \textbf{REFID}             & \cmark & \cmark  & \textbf{35.90} & \textbf{0.973} &  & \textbf{35.47} & \textbf{0.971} & 15.9M   \\ \hline
    
    \textbf{HighREV}   &   &         & \multicolumn{2}{c}{\textbf{11+1}}       &  & \multicolumn{2}{c}{\textbf{11+3}} &     \\ \hline
    \Add{\textbf{REFID (\textit{ssl.})}} & \cmark & \cmark  & \Add{\textbf{32.01}}  &  \Add{\textbf{0.891}}  &  & \Add{\textbf{31.16}} & \Add{\textbf{0.881}} &  \Add{\textbf{7.8M}}   \\
    RIFE~\cite{huang2020rife}         & \cmark & \xmark  & 32.79 & 0.904 &  & 31.24 & 0.890 & 9.8M \\
    EDI~\cite{edi_pan}                & \cmark & \cmark  & 24.48 & 0.735 &  & 23.53 & 0.715  & 0.5M \\
    EFNet+IFRNet~\cite{sun2022event,ifrnet} & \cmark & \cmark  & 35.97  & 0.959  &  & 35.42 & 0.966  & 28.2M     \\
    \revise{\textbf{REFID (\textit{ssl. fine-tuning})}} & \cmark & \cmark  & \revise{\textbf{36.05}}  &  \revise{\textbf{0.961}}  &  & \revise{\textbf{35.51}} & \revise{\textbf{0.966}} &  \revise{\textbf{7.8M}}   \\
    E2VID+~\cite{rebecq2019events}   & \cmark & \cmark  & 36.36  & 0.970  &  & 35.77 & 0.968  & 15.3M \\
    \textbf{REFID}            & \cmark & \cmark  & \textbf{37.65}  &  \textbf{0.975}  &  & \textbf{36.91} & \textbf{0.973} & 15.9M     \\ \hline
    \end{tabular}}
    \label{tab:deblurinterpo}
\end{table*}

\begin{table}[!t]
\caption{{Comparison of single image motion deblurring methods on GoPro~\cite{nah2017deep} and REBlur~\cite{sun2022event}.} HINet+: event-enhanced versions of HINet~\cite{chen2021hinet}. \Add{\textit{Ssl.} denotes that the model is self-supervised trained from scratch without ground-truth as supervision.}}
\centering
\small
\resizebox{0.48\textwidth}{!}{
\setlength{\tabcolsep}{10pt}
\renewcommand\arraystretch{1.0}
\begin{tabular}{ l c  c  c  c }
\bottomrule[0.15em]
\rowcolor{tableHeadGray}
\textbf{Method} &\textbf{Events} & \textbf{PSNR} $\uparrow$ & \textbf{SSIM} $\uparrow$ & \textbf{\#Param} \\ \hline \hline
\textbf{GoPro}~\cite{nah2017deep} \\ \hline
EDI~\cite{edi_pan}                  &\cmark & 27.34 & 0.901 \\
\Add{\textbf{REFID (\textit{ssl.})}}         & \Add{\cmark}   & \Add{\textbf{28.88}} & \Add{\textbf{0.912}} & \Add{\textbf{7.8M}} \\
D$^{2}$Nets$^{\dagger}$~\cite{shang2021bringing} &\cmark      & 31.60  & 0.940 & - \\
LEMD$^{\dagger}$~\cite{jiang2020learning}    & \cmark         & 31.79  & 0.949 & - \\
MPRNet~\cite{zamir2021multi}                 & \xmark         & 32.66  & 0.959 & 20.0M \\
Restormer~\cite{Zamir2021Restormer}          & \xmark         & 32.92  & 0.961 & 26.1M \\
ERDNet~\cite{haoyu2020learning}              & \cmark         & 32.99  & 0.935 & - \\
NAFNet~\cite{chen2022simple}                 & \xmark         & 33.69  & 0.967 &  - \\
EFNet~\cite{sun2022event}                    & \cmark         & 35.46 & 0.972  & 8.5M \\
\textbf{REFID}                        & \cmark         & \textbf{35.91} & \textbf{0.973} & 15.9M \\

\hline
\textbf{REBlur~\cite{sun2022event}} \\ \hline
\Add{\textbf{REFID (\textit{ssl.})}}         & \Add{\cmark}  & \Add{\textbf{35.01}} & \Add{\textbf{0.953}}  & \Add{\textbf{7.8M}} \\
SRN~\cite{tao2018scale}                      & \xmark    & 35.10 & 0.961  & 10.3M \\
NAFNet~\cite{chen2022simple}                 & \xmark    & 35.48 & 0.962  & 67.9M \\
Restormer~\cite{Zamir2021Restormer}          & \xmark    & 35.50 & 0.959  & 26.1M \\
EDI~\cite{edi_pan}                           & \cmark    & 36.52 & 0.964  & 0.5M \\
HINet+~\cite{chen2021hinet}                  & \cmark    & 37.68 & 0.973  & 88.9M \\ 
EFNet~\cite{sun2022event}                    & \cmark    & {38.12} & \textbf{0.975} & 8.5M \\  
\textbf{REFID}                               & \cmark    & \textbf{38.34} & \textbf{0.975} & 15.9M \\

\hline
\end{tabular}
}
\label{table:deblur}
\end{table}

\subsection{Sharp Frame Interpolation}
\label{subsec:sharpFI}
% We report sharp frame interpolation results in Tab.~\ref{tab:sharpinterpo}. Our method achieves state-of-the-art performance in the 7- and 15-skip settings on both examined datasets, improving upon competing methods substantially \Add{(1.99 dB/2.43 dB and 5.99 dB/9.36 dB improvement compared with state-of-the-art on GoPro and HighREV, respectively).} Fig.~\ref{fig:sharp_interpo} shows qualitative results on HighREV. RIFE shows artifacts because of the ambiguity of motion in the time between the two images. Our method exhibits stable performance both on indoor and outdoor scenes.
\Add{We present the results of sharp frame interpolation in Table~\ref{tab:sharpinterpo}. Our proposed method demonstrates state-of-the-art performance in the 7- and 15-skip settings across both examined datasets, manifesting substantial improvements over competing methods. Specifically, our approach achieves a notable enhancement of 1.99 dB and 2.43 dB in the GoPro dataset and 5.99 dB and 9.36 dB in the HighREV dataset, respectively, compared to the current state-of-the-art. Qualitative results on HighREV, illustrated in Figure~\ref{fig:sharp_interpo}, highlight the efficacy of our method. Notably, RIFE exhibits artifacts attributable to the ambiguity of motion between the two images, while our method maintains stable performance across diverse scenes, encompassing both indoor and outdoor environments.}

\revise{Experiments on the BS-ERGB~\cite{tulyakov2022time} dataset please refer to the Supplementary Materials.}

\subsection{Blurry Frame Interpolation}
\label{subsec:blurryFI}
We compare our method with state-of-the-art image-only and event-based methods. Since most event-based methods do not have public implementations, we use ``E2VID+'' by adding an extra encoder for images and introduce images as extra inputs for the event-based image reconstruction method E2VID~\cite{Rebecq19cvpr}. As a two-stage method, we use EFNet+IFRNet by combining a state-of-the-art event-based image deblurring method~\cite{sun2022event} with an image-only frame interpolation method~\cite{ifrnet}. For a fair comparison, IFRNet is also fed with event voxels from two directions as inputs. For Time Lens~\cite{tulyakov2021time}, because the training code is not available, we use the public model and pre-trained weights.

\begin{figure*}[!tbp]
    \centering 
    \includegraphics[width=0.95\textwidth]{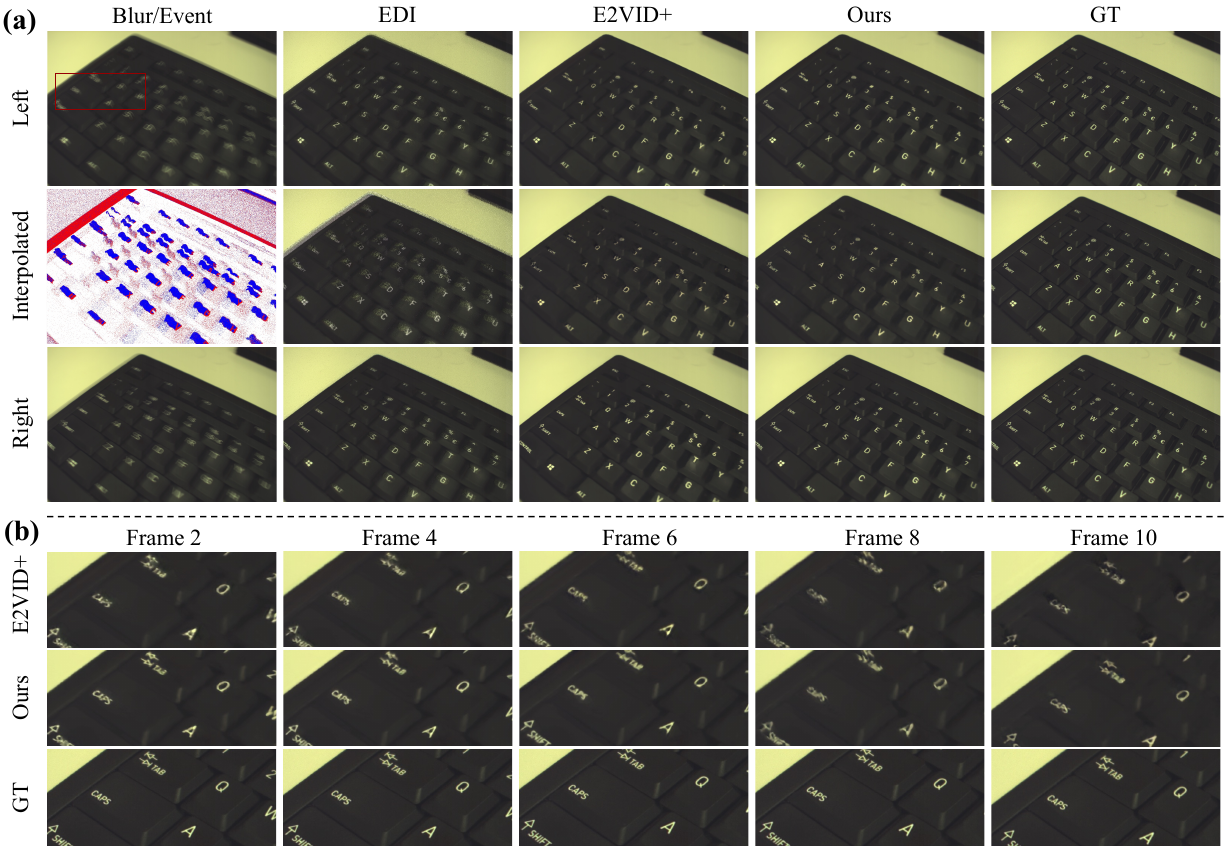}
    \caption{{(a): Visual comparison on HighREV of the restored left, right and interpolated frame.} E2VID+: image-enhanced version of E2VID~\cite{Rebecq19cvpr}. Compared to other event-based methods, our method achieves the most faithful results. {(b): The interpolated frames in the exposure time of the left blurry image.} Best viewed on a screen and zoomed in.}
    \label{fig:blur_interpo}
\end{figure*}

Quantitative results are reported in Tab.~\ref{tab:deblurinterpo}. Although our method can also interpolate latent frames in the exposure time, the results are reported on the interpolated frame between the two exposure times.
\ourmodel achieves 2.08 dB/0.012 and 1.29 dB/0.005 improvement in PSNR and SSIM on the ``11+1'' setting on GoPro and HighREV, respectively. For the ``11+3'' setting, the improvements over the second-best method amount to 2.08 dB/0.017 and 1.14 dB/0.005, showing that our principled bidirectional architecture with event-guided fusion leverages events more effectively. \Add{Even in the absence of ground truth for training (i.e., the \textit{ssl.} version of REFID), our method surpasses EDI and Time Lens, underscoring the efficacy of the proposed self-supervised training framework.}\revise{When applying supervised pre-training on the synthetic dataset (GoPro) and adapting to real-world dataset with the proposed self-supevisde framework, the result increases from 32.01 dB to 36.06 dB in the ``11+1'' setting.}. The state-of-the-art event-based method Time Lens exhibits a large performance degradation on blurry frame interpolation because of the assumption of sharp keyframes and neglecting the intensity changes within the exposure time.
Fig.~\ref{fig:blur_interpo} shows qualitative results. Fig.~\ref{fig:blur_interpo} (a) depicts the results for the left, right and interpolated frame on HighREV. EDI~\cite{edi_pan} is vulnerable to noise and inaccurate events. E2VID+ exhibits artifacts because its unidirectional architecture does not leverage future events. \ourmodel achieves sharp and faithful results both on textured regions and edges thanks to the bidirectional architecture and its event-guided attention fusion. Fig.\ref{fig:blur_interpo} (b) shows the results of frame interpolation within the exposure time.

% The self-supervised training framework proposed in our paper is tailored for our REFID framework. Deblurring an image to an image sequence is an integral part of our self-supervised framework because all the loss functions (Brightness Increment Loss, Blur Consistency Loss, and Warping Loss) are built on the assumption that a sequence of images is predicted corresponding to a single blurry image. Thus, single-image deblurring methods like EFNet are not applicable to our self-supervised framework.

\subsection{Single Image Deblurring}
\label{subsec:deblur}
As a by-product, REFID can also perform single-image motion deblurring, and Tab.~\ref{table:deblur} reports quantitative comparisons on this task. 
\revise{It is worth mentioning that REFID predicts a short sequence of images that are within the exposure time of the input blurry image. And this is different from other traditional single-image deblurring methods. Thus, single-image deblurring results from our REFID represent the averaged PSNR and SSIM from the resulting image sequences.}

With the supervised-training strategy, compared with the state-of-the-art EFNet~\cite{sun2022event}, our method pushes the performance further to 35.91 dB in PSNR on GoPro. The 0.22 dB improvement in PSNR over EFNet on REBlur also evidences the robustness of REFID on real-world blurry scenes.

\Add{
% add ssl training.
Considering a more realistic scene, the ground-truth sharp images for most of the event cameras are not available or hard to get. In these conditions, the models trained on synthetic data are used for real conditions. SRN+, HINet+, EFNet, and REFID in Tab.\ref{table:ssl} are examples. 
For our self-supervised training framework, we can choose either training from scratch on the real-world dataset (HighREV dataset) or self-supervised fine-tuning with pre-trained weights from synthetic datasets, with 1.32 dB and 3.98 dB higher than state-of-the-art.}

\begin{figure*}
    \centering
    % \vspace{-3ex}
    \includegraphics[width=0.98\textwidth]{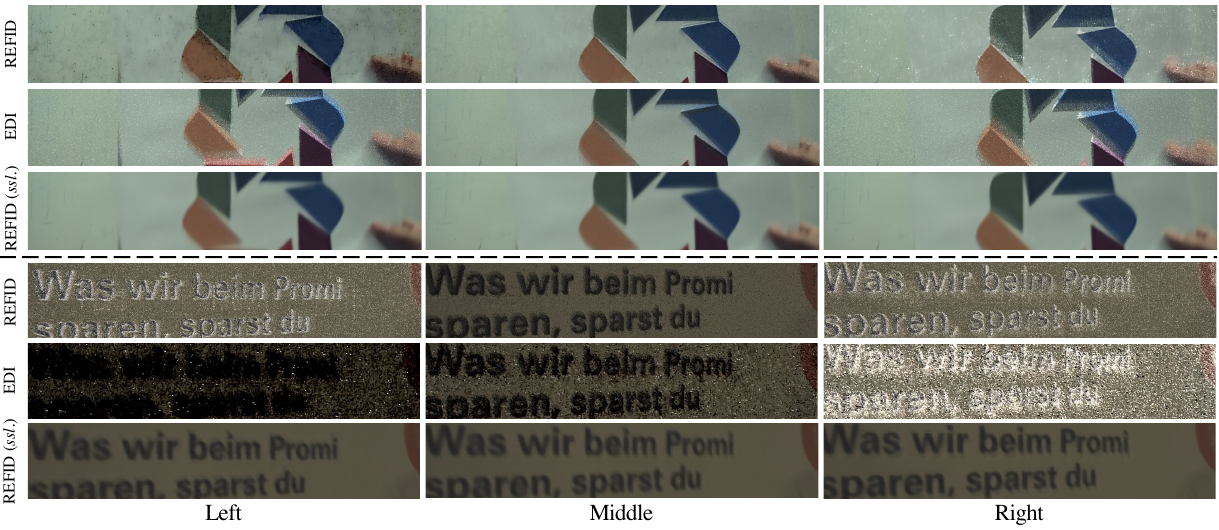}
    \caption{{\textbf{\revise{Qualitative results of the single-image deblurring without ground-truth.}} \revise{The terms ``Left'', ``Middle'', and ``Right'' denote the first, middle, and last latent sharp images within the exposure time, respectively. Our method, incorporating a self-supervised framework, effectively mitigates noise accumulation in the events, ensuring consistent and high-quality deblurring results in the generated video clip.}}}
    \label{fig:ssl_interpo}
    % \vspace{-3ex}
\end{figure*}

\Add{
Because of the domain gap between the synthetic data and realistic data, the performance of the models is downgraded. We choose REFID as a representative method and show the qualitative results in Fig.\ref{fig:ssl_interpo}. Although the result in the middle of the exposure time is good, the other results in the exposure time are vulnerable to the influence of accumulated noise. The ``Left'' and ``Right'' results in Fig.\ref{fig:ssl_interpo} shows the artifacts.
The settings for EDI are similar to our method because it is also a self-supervised method, but it fails to deal with the spatially-variant contrast threshold, leading to artifacts in the blurry areas.
Equipped with our proposed self-supervised training framework (denoted as \textit{ssl.} in Tab.~\ref{table:deblur}), our method utilizes predicted optical flow as a substitute for events, getting rid of the influence of accumulated noise. The qualitative results show the strong generalization of the proposed framework on real-world data.}
\revise{Note that because our proposed self-supervised framework is built on the assumption that the model predicts a short video clip with the timestamps within the exposure time of the input image, traditional single-image deblurring methods like EFNet is not applicable to our method.}

\begin{table}[t]
\caption{\Add{{Comparison of event-based motion deblurring methods on the HighREV dataset.} Methods with a + denote event-enhanced versions.}}
\centering
\small
\resizebox{0.48\textwidth}{!}{
\setlength{\tabcolsep}{6pt}
\renewcommand\arraystretch{1.0}
\begin{tabular}{ l  || c |  c  |  c |  c }
\bottomrule[0.15em]
\rowcolor{tableHeadGray}
  & \textbf{sup. train.} & \textbf{self sup. train.} &  & \\
 \rowcolor{tableHeadGray}
 \multirow{-2}{*}{\textbf{Method}}&\textbf{on GoPro}&\textbf{on HighREV}&\multirow{-2}{*}{\textbf{PSNR}$\uparrow$}&\multirow{-2}{*}{\textbf{SSIM}$\uparrow$}\\
 
 \hline \hline
EDI~\cite{edi_pan}             & \xmark   & \cmark        & 25.32            & 0.753 \\
% SRN~\cite{tao2018scale}        & \xmark           & -                & -       \\
SRN+~\cite{tao2018scale}       & \cmark   & \xmark        & 25.65            & 0.836  \\
% HINet~\cite{chen2021hinet}     & \xmark           & -                & -       \\
HINet+~\cite{chen2021hinet}    & \cmark   & \xmark        & 28.70            & 0.910    \\
EFNet~\cite{sun2022event}      & \cmark   & \xmark        & 29.55            & 0.936   \\
% REFID~\cite{sun2023event}      & \cmark   & \xmark        & 28.97            & 0.894  \\\hline
\textbf{REFID}                           & \cmark   & \xmark        & 28.72            & 0.910 \\
\textbf{REFID (\textit{ssl.})}                           & \xmark   & \cmark        & 30.04            & 0.931 \\
\textbf{REFID (\textit{pre-trained ssl.})}               & \cmark   & \cmark        & 32.70            & 0.951  \\
% \bottomrule
\bottomrule
\end{tabular}
}
\label{table:ssl}
\end{table}

\begin{table}[t]
    \caption{{Ablation study of different architectural components of our method} on the GoPro~\cite{nah2017deep} dataset using the ``11+1'' setting.}
    \resizebox{0.48\textwidth}{!}{%
    \setlength{\tabcolsep}{6pt}
    \renewcommand\arraystretch{1.1}
    \begin{tabular}{c|c|c||c|c}
    \bottomrule[0.15em]
    \rowcolor{tableHeadGray}
    \textbf{Multi-scale connection}  & \textbf{Fusion}   &  \textbf{Recurrent}   
    & \textbf{PSNR} & \textbf{SSIM}  \\ \hline \hline
     \xmark                 & add         &  \xmark    &    33.24  & 0.950 \\
     \cmark                 & add         &  \xmark    &    33.61  & 0.952 \\ 
     \cmark                 & add         &  ConvLSTM  &    34.39  & 0.962\\ 
     \cmark                 & add         &  ConvGRU    &    34.54  & 0.962\\ 
     \cmark                 & add         &  EVR unidir.  &    35.36  & 0.968\\ 
     \cmark                 & add         &  EVR bidir.  &    35.81  & 0.971\\ 
     \cmark                 & EGACA       &  EVR bidir.  &   \textbf{36.12} & \textbf{0.974} \\ \hline
    \end{tabular}
    }
    \label{tab:ablation_architecture}
\end{table}

\subsection{Ablation Study}
\label{subsec:ablation}
\Add{\textbf{Supervised training:}} \Add{Ablation studies on supervised training are conducted on GoPro with the ``11+1'' setting to analyze the effectiveness of the proposed model architecture and its components (Tab.~\ref{tab:ablation_architecture})}. First, the proposed recurrent architecture improves PSNR by 1.75 dB compared to the non-recurrent architecture, proving the effectiveness of temporal modeling of events. Furthermore, the proposed bidirectional EVR block yields an improvement of 0.45 dB in PSNR compared to its unidirectional counterpart, showcasing the informativeness of future events and the merit of our physically-based model design. Compared to ConvLSTM~\cite{shi2015convolutional} and ConvGRU~\cite{shi2017deep}, which model longer time dependencies and are used in video recognition, our EVR block using a simple ResNet~\cite{he2016resnet} yields 0.84 dB improvement in PSNR.
Moreover, the proposed EGACA contributes an improvement of 0.31 dB, evidencing the benefit of mining and fusing image features with adaptive weights from current events.
The multi-scale connection between the image branch and the event branch also brings a 0.37 dB gain in PSNR.
Finally, all our contributions together yield a substantial improvement of 2.88 dB in PSNR and 0.024 in SSIM over the baseline.

{\PAR{Self-supervised training:}}
\Add{Ablation studies in Tab.~\ref{table:ablation_loss} and Tab.~\ref{table:ablation_module} are conducted with self-supervised training from scratch on each dataset.}

\Add{Tab.~\ref{table:ablation_loss} shows the distinct contributions of different loss components — brightness increment loss, blur consistency loss, and warp loss — to the performance of our image deblurring model, and reports PSNR and SSIM on GoPro and HighREV.
The blur consistency loss appears critical for achieving decent performance, as reflected by the results where this loss was excluded (i.e., the combination of brightness increment loss and warp loss). The model performance dropped dramatically on both datasets, reaching a PSNR of only 7.22 dB and 10.26 dB on GoPro and HighREV respectively, with correspondingly low SSIM values.
On the other hand, the inclusion of the warp loss is highly beneficial. For instance, when this loss was added to the blur consistency loss, the PSNR on the HighREV dataset increased substantially from 23.41 dB to 28.53 dB, with similar improvements on the synthetic GoPro dataset.
The addition of the brightness increment loss to the blur loss led to only minor improvements on the synthetic GoPro but substantially benefits the real-world results on the HighREV dataset where the PSNR increases from 23.41 dB to 27.81 dB. This difference in performance boost could stem from the fact that the events in the GoPro dataset are generated with interpolated frames that often yield subtle artifacts that affect synthetic event generation.
Finally, it is worth noting that the combination of all three losses yielded the best results. The model achieved a PSNR of 28.88 dB on GoPro and 30.04 dB on HighREV, with high SSIM values, demonstrating the complementary roles of these loss components.
This finding reinforces the effectiveness of our proposed framework that capitalizes on the different constraints offered by each of these losses in the image deblurring process.}
%\begin{figure}
%    \centering
%    \includegraphics[width=0.98\textwidth]{figures/ablation.pdf}
%    \caption{\textbf{Visualized examples for ablation study.}}
%    \label{fig:ablation}
%\end{figure}

\Add{Next we investigate the use of images of warped events at the input of our method, and the use of RobustNorm+, a voxel normalization method that deviates slightly from the RobustNorm in~\cite{stoffregen2020reducing}. 
Both normalize non-zero values as
\begin{equation}
    \begin{aligned}
    \bar{\mathbf{V}}^{\text{RN}}_i(\textbf{x}) &= \text{clip}\left(\frac{\mathbf{V}_i(\textbf{x}) - \textbf{V}_{i,1}}{\textbf{V}_{i,99} - \textbf{V}_{i,1}}, 0, 1\right), \\
    \bar{\mathbf{V}}^{\text{RN+}}_i(\textbf{x}) &= \text{clip}\left(\frac{\mathbf{V}_i(\textbf{x})}{\textbf{V}_{i,99}}, 0, 1\right), \\
    \end{aligned}
\end{equation}
where $\textbf{V}_{i,p}$ denotes the $p^\text{th}$ percentile of $\mathbf{V}_i(\textbf{x})$. We found that RobustNorm+ was more stable during training and lead to fewer outliers on samples with few events. We report an ablation of these components in Tab.~\ref{table:ablation_module}. Removing IWE from the input led to a 0.19 dB reduction on GoPro, and a 0.90 dB reduction on HighREV, justifying its use. We argue that the IWE is essential since it provides a sharp template for the network for placing edges at the output. We also find that RobustNorm+ increases the PSNR by 0.89 dB on GoPro and 1.69 dB on HighREV.}

\begin{table}[!t]
    \caption{\Add{{Ablation on losses.} Brightness increment ($\mathcal{L}_\text{br. inc.}$), blur consistency ($\mathcal{L}_\text{blur cons.}$), and warping loss ($\mathcal{L}_\text{warp}$).}}
    \resizebox{0.48\textwidth}{!}{%
    \setlength{\tabcolsep}{6pt}
    \renewcommand\arraystretch{1.1}
    \begin{tabular}{ c  c  c || c   c  |  c  c }
    \bottomrule[0.15em]
    \rowcolor{tableHeadGray}
    & &  & \multicolumn{2}{c|}{\textbf{GoPro}} & \multicolumn{2}{c}{\textbf{HighREV}} \\
    %\cline{4-7}
    \rowcolor{tableHeadGray}
    \multirow{-2}{*}{\textbf{$\mathcal{L}_\text{br. inc.}$}} & \multirow{-2}{*}{\textbf{$\mathcal{L}_\text{blur cons.}$}} & \multirow{-2}{*}{\textbf{$\mathcal{L}_\text{warp}$}} & \textbf{PSNR} & \textbf{SSIM} & \textbf{PSNR} & \textbf{SSIM} \\ \hline \hline
    %\cmark & \xmark & \xmark & 6.84 & 0.044 & 9.425 & 0.131 \\
    \xmark & \cmark & \xmark & 23.52 & 0.824 & 23.41 & 0.817 \\\hline
    %\xmark & \xmark & \cmark & 6.38 & 0.056 & 9.77 & 0.130 \\ \hline
    \cmark & \cmark & \xmark & 23.78 & 0.831 & 27.81 & 0.925 \\
    \cmark & \xmark & \cmark & 7.22 & 0.026 & 10.26 & 0.175 \\
    \xmark & \cmark & \cmark & 27.72 & 0.893 & 28.53 & 0.927 \\ \hline
    \cmark & \cmark & \cmark & 28.88 & 0.912 & 30.04 & 0.931 \\
    % \bottomrule
    \hline
    \end{tabular}
    }
    \label{table:ablation_loss}
\end{table}

\begin{table}[!t]
    \caption{\Add{{Ablation on components in self-supervised learning framework.} IWE: Using images of warped events for training. Voxel Norm: the method used for voxel normalization.}}
    \resizebox{0.48\textwidth}{!}{%
    \setlength{\tabcolsep}{6pt}
    \renewcommand\arraystretch{1.1}
    \begin{tabular}{ c  c || c   c  |  c  c }
    \bottomrule[0.15em]
    \rowcolor{tableHeadGray}
    &  & \multicolumn{2}{c|}{\textbf{GoPro}} & \multicolumn{2}{c}{\textbf{HighREV}} \\
    %\cline{4-7}
    \rowcolor{tableHeadGray}
    \multirow{-2}{*}{\textbf{IWE}} & \multirow{-2}{*}{\textbf{Voxel Norm}} & \textbf{PSNR} & \textbf{SSIM} & \textbf{PSNR} & \textbf{SSIM} \\ \hline \hline
    \xmark & RobustNorm~\cite{Stoffregen19cvpr} & 27.95 & 0.896 & 28.98 & 0.894 \\
    \cmark & RobustNorm~\cite{Stoffregen19cvpr} & 27.99 & 0.895 & 29.25 & 0.924 \\ 
    \xmark & RobustNorm+                               & 28.59 & 0.909 & 30.04 & 0.931 \\\hline
    \cmark & RobustNorm+                               & \textbf{28.88} & \textbf{0.912} & \textbf{30.94} & \textbf{0.937} \\
    % \bottomrule
    
    \hline
    \end{tabular}
    }
    \label{table:ablation_module}
\end{table}

\section{Conclusion}
\label{sec:conclusion}

In this paper, we have considered the tasks of event-based sharp frame interpolation and blurry frame interpolation jointly, as motion blur may or may not occur in input videos depending on the speed of the motion and the length of the exposure time. To solve these tasks with a single method, we have proposed \ourmodel, a novel bidirectional recurrent neural network which performs fusion of the reference video frames and the corresponding event stream. The recurrent structure of \ourmodel allows the effective propagation of event-based information across time, which is crucial for accurate interpolation. Moreover, we have introduced \ourfusionmodule, a new adaptive event-image fusion module based on channel attention. In order to provide a more realistic experimental setting for the examined low-level event-based tasks, we have presented \ourdataset, a new event-RGB dataset with the highest spatial event resolution among related sets. We have thoroughly evaluated our network on standard event-based sharp frame interpolation, event-based blurry frame interpolation, and single-image deblurring and shown that it consistently outperforms existing state-of-the-art methods on GoPro and \ourdataset.
\Add{To improve the generalization of the model, we propose a self-supervised training framework with warped events for blurry frame interpolation and single-image deblurring. Three loss functions are utilized for the proposed framework. Experiments on the real-world dataset are conducted to show the effectiveness of the framework. We hope the work can inspire more event-based computational imaging work for realistic applications.}

\Add{In pursuit of enhancing the model's generalization, we further introduce a self-supervised training framework incorporating warped events for both blurry frame interpolation and single-image deblurring. The proposed framework integrates three distinct loss functions to constrain the model training. Through comprehensive experiments conducted on real-world datasets, we substantiate the effectiveness of our proposed approach. We envision that this work will serve as a catalyst for inspiring further exploration in the domain of event-based computational imaging for realistic applications.}

% \PAR{Acknowledgments.}
% This work was supported by the National Natural Science Foundation of China (NSFC) under 
% Grant No.12174341, the National Key R\&D Program of China under Grant No.2022YFF0705500 
% and No.2022YFB3206000, the China Scholarship Council, and AlpsenTek GmbH.

\ifCLASSOPTIONcaptionsoff
  \newpage
\fi

%%%%%%%%% REFERENCES
{\small
\bibliographystyle{ieee_fullname}
\bibliography{egbib}
}

%%%%%%%%%% BIO %%%%%%%%%%%
% \newpage

 

% \newpage

\begin{IEEEbiography}[{\includegraphics[width=1in,height=1.25in,clip,keepaspectratio]{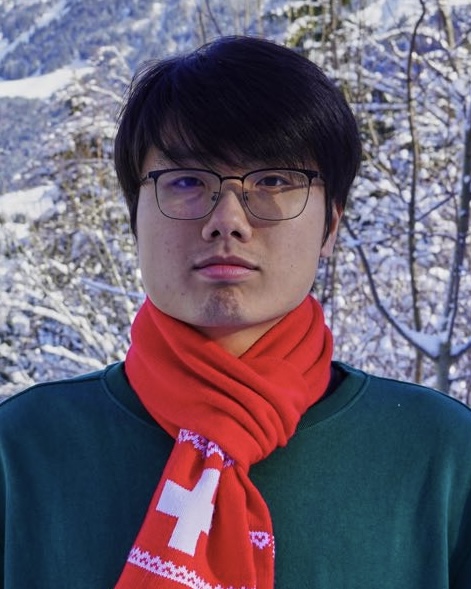}}]Lei Sun received his Ph.D. from Zhejiang University in 2024 and his B.S. from Beijing Institute of Technology in 2018. He is currently a postdoctoral researcher at INSAIT, Sofia University ``St. Kliment Ohridski'', Bulgaria. He was a recipient of the Xiaomi Special Scholarship and the CSC Scholarship. During his doctoral studies, he was a visiting researcher at the Computer Vision Lab, ETH Zürich, and the Robotics and Perception Group, University of Zürich. His research interests include event-based vision, low-level vision, and video generation. For more information, visit his website: https://ahupujr.github.io/.
\end{IEEEbiography}
\vskip -1\baselineskip plus -1fil

\begin{IEEEbiography}
[{\includegraphics[width=1in,height=1.25in,clip,keepaspectratio]{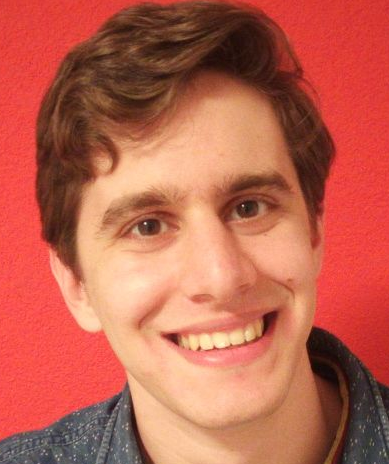}}]{Daniel Gehrig} obtained his Ph.D. with the highest distinction from the Robotics and Perception Group (RPG), at the University of Zurich, Switzerland for which he was granted the UZH Annual Award. Between 2012 to 2018 he completed his master’s studies in Mechanical Engineering at ETH Zurich, with the highest possible score, for which he was awarded the Willi Studer Prize and ETH Medal for the best master’s thesis of the year. His research interests lie at the intersection of robotics, computer vision, and machine learning for event-based vision. His work has been featured prominently in IEEE Spectrum and on popular channels like Two Minute Papers.
\end{IEEEbiography}
\vskip -1\baselineskip plus -1fil

\begin{IEEEbiography}[{\includegraphics[width=1in,height=1.25in,clip,keepaspectratio]{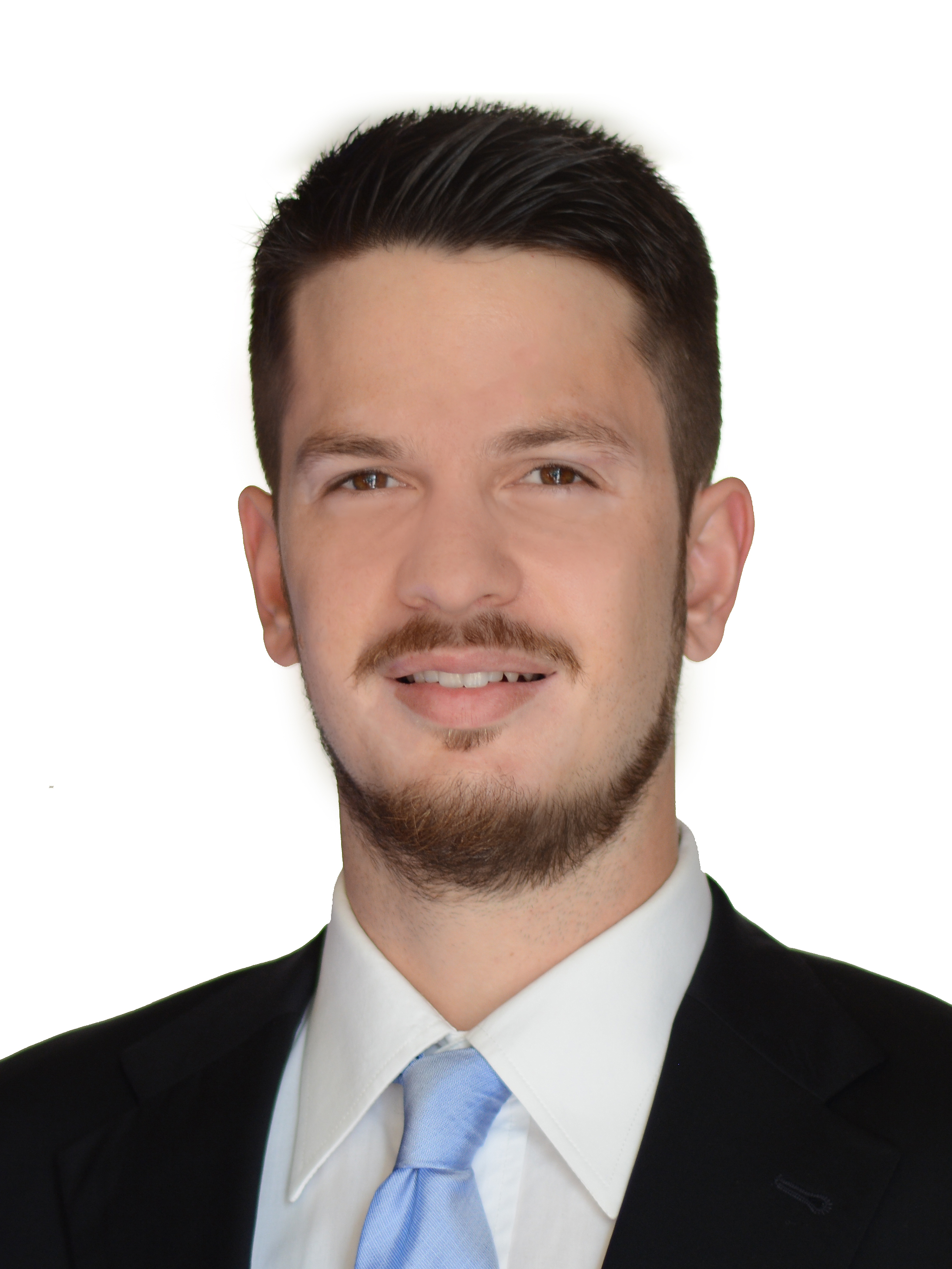}}]{Christos Sakaridis} is a lecturer at ETH Zurich and a senior postdoctoral researcher at the Computer Vision Lab of ETH Zurich. His research fields are computer vision and machine learning. The focus of his research is on semantic and geometric visual perception, involving multiple domains, visual conditions, and visual or non-visual modalities. Since 2021, he is the Principal Engineer in TRACE-Zurich, a large-scale project on computer vision for autonomous cars and robots. He received the ETH Zurich Career Seed Award in 2022. He obtained his PhD from ETH Zurich in 2021, having worked in Computer Vision Lab. Prior to that, he received his MSc in Computer Science from ETH Zurich in 2016 and his Diploma in Electrical and Computer Engineering from National Technical University of Athens in 2014.
\end{IEEEbiography}
\vskip -1\baselineskip plus -1fil

\begin{IEEEbiography}[{\includegraphics[width=1in,height=1.25in,clip,keepaspectratio]{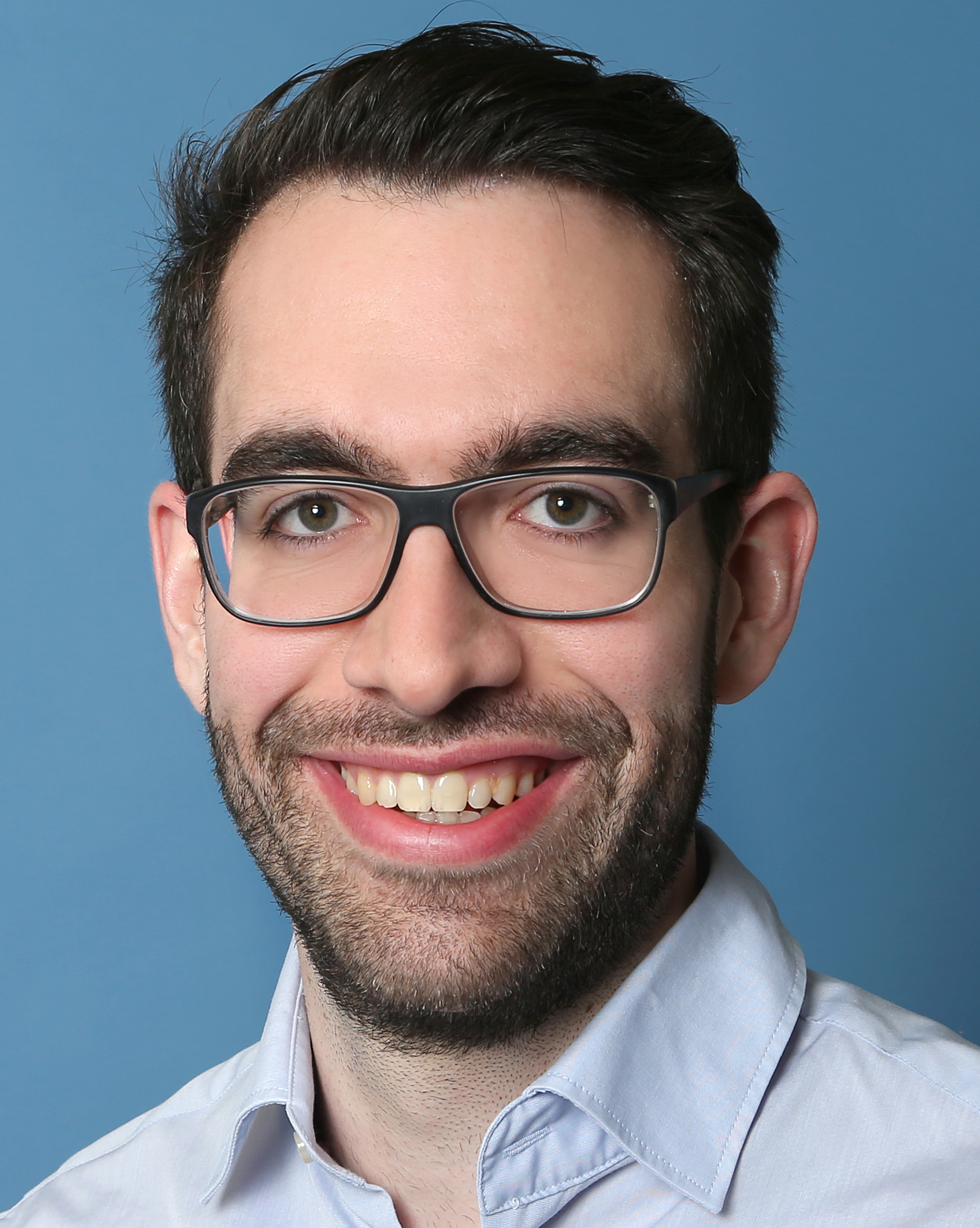}}]{Mathias Gehrig} %
%obtained his M.Sc. degree in Robotics, Systems and Control at ETH Zürich, Switzerland in 2016.
%Previously, he received a B.Sc. degree in Mechanical Engineering (2013).
%He is a Ph.D. candidate in computer science at the University of Zürich under the supervision of Prof. Davide Scaramuzza.
%He is broadly interested in the application of machine learning for real-time computer vision and robotics.
obtained his M.Sc. in Robotics, Systems and Control from ETH Zurich, Switzerland, in 2016, after receiving his B.Sc. in Mechanical Engineering in 2013. As a Ph.D. candidate in computer science at the University of Zurich, supervised by Prof. Davide Scaramuzza, he focuses on the application of machine learning for real-time computer vision and robotics.
Notably, his work was nominated for the Best Paper Award at the 2023 Conference on Computer Vision and Pattern Recognition (CVPR).
\end{IEEEbiography}
\vskip -1\baselineskip plus -1fil

\begin{IEEEbiography}[{\includegraphics[width=1in,height=1.25in,clip,keepaspectratio]{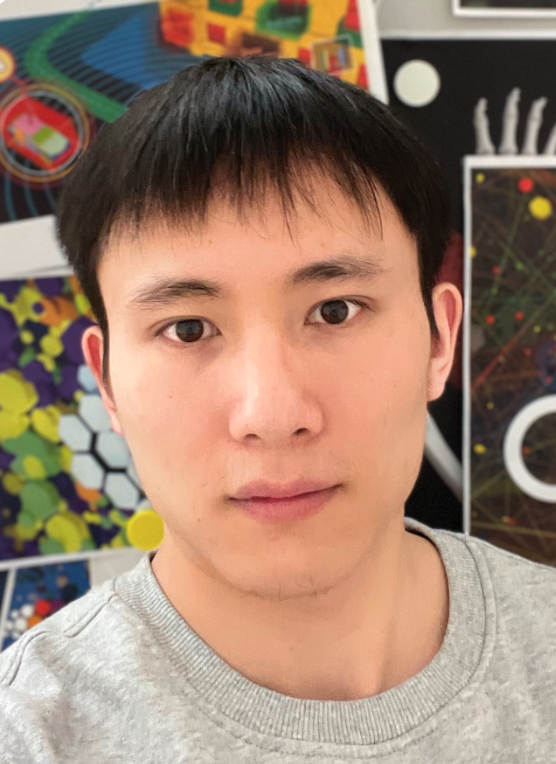}}]{Jingyun Liang} is currently a  PhD student at Computer Vision Lab, ETH Zurich. He received his B.S degree and master's degree from National University of Defense Technology in 2014 and 2016, His research focuses on low-level vision research, especially on image and video restoration, such as super-resolution, deblurring and denoising.
\end{IEEEbiography}
\vskip -1\baselineskip plus -1fil

\begin{IEEEbiography}[{\includegraphics[width=1in,height=1.25in,clip,keepaspectratio]{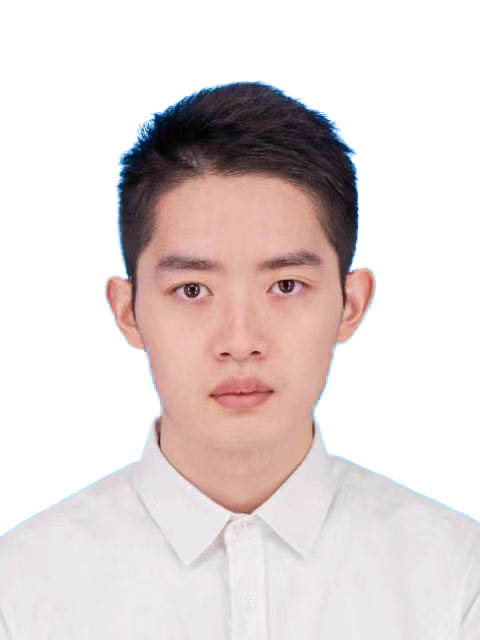}}]{Peng Sun} received his B.S degree in Optical Engineering from Zhejiang University(ZJU) in 2020, and gain his master degree in information engineering in 2023 from the National Research Center, Zhejiang University. His research focuses on event-based vision, image deblurring, and visual detection algorithms.
\end{IEEEbiography}
\vskip -1\baselineskip plus -1fil

\begin{IEEEbiography}[{\includegraphics[width=1in,height=1.25in,clip,keepaspectratio]{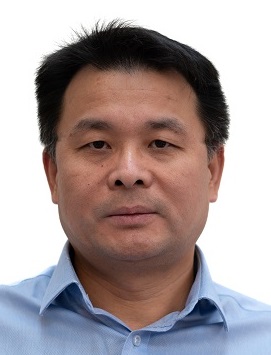}}]{Zhijie Xu}, received his PhD from the University of Derby, UK, in 2000. He is a senior academic director of Graduate Education and director of the Centre for Visual and Immersive Computing at the University of Huddersfield
He has published 200 peer-reviewed papers and edited five books. He has mentored 60 postgraduates, including 15 PhD students to completion. He has served as General Chair for conferences, including the 23rd IEEE ICAC.
\end{IEEEbiography}
\vskip -1\baselineskip plus -1fil

\begin{IEEEbiography}[{\includegraphics[width=1in,height=1.25in,clip,keepaspectratio]{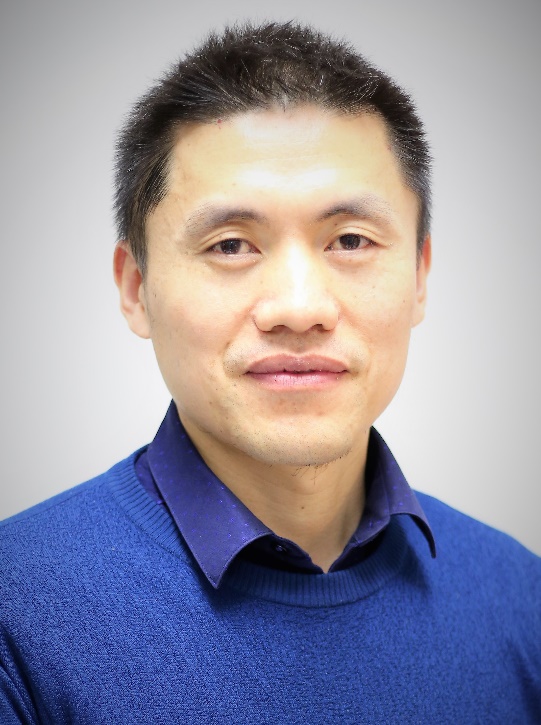}}]{Kaiwei Wang} is currently a full professor at the State Key Laboratory of Modern Optical Instrumentation, and the Deputy Director of the National Optical Instrument Engineering Research Center at Zhejiang University. He received a B.S. degree in 2001 and a Ph.D. degree in 2005 respectively, both from Tsinghua University. In October 2005, he started his postdoctoral research at the Center of Precision Technologies (CPT) of Huddersfield University, funded by the Royal Society International Visiting Postdoctoral Fellowship and the British Engineering Physics Council. He joined Zhejiang University in February 2009 and has been mainly researching on intelligent optical sensing technology and visual assisting technology for the visually impaired. Up to date, he owns 80 patents and has published more than 150 refereed research papers. For more information, visit his Website: http://wangkaiwei.org/.
\end{IEEEbiography}
\vskip -1\baselineskip plus -1fil

\begin{IEEEbiography}[{\includegraphics[width=1in,height=1.25in,clip,keepaspectratio]{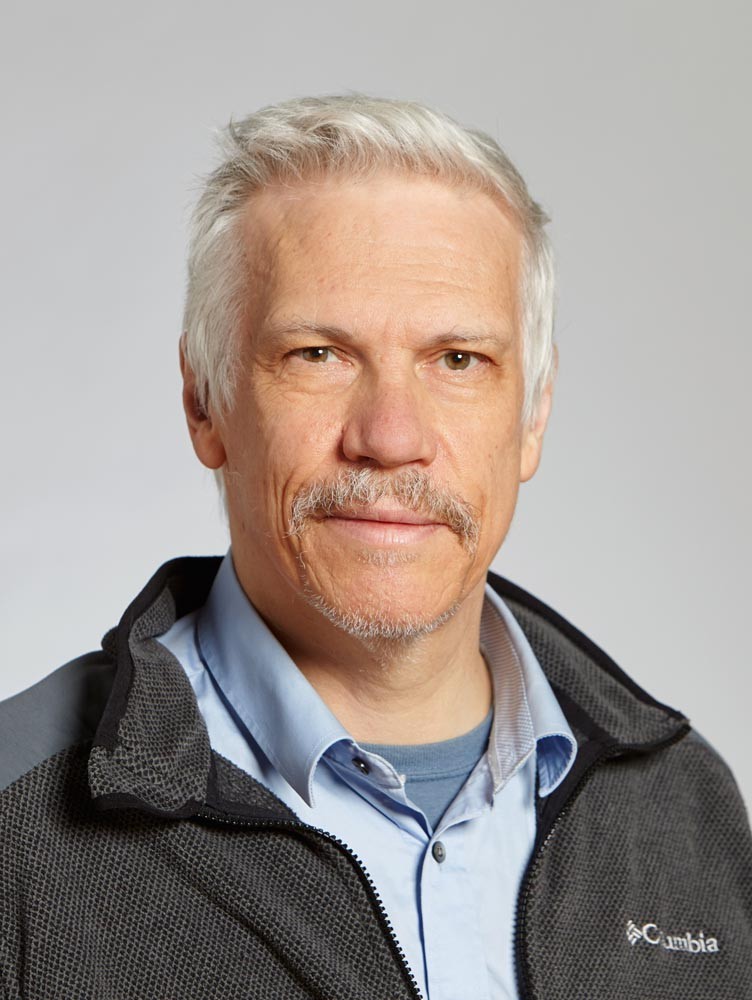}}]{Luc Van Gool} is both a full professor with KU Leuven (Belgium) and with ETH Zurich (Switzerland). His main area of expertise is computer vision. He received the Koenderink Prize with the European Conference on Computer Vision, in 2016, the David Marr prize (best paper award) at the International Conference on Computer Vision, in 1998 and the U.V. Helava Award, one of the most prestigious ISPRS awards, in 2012. He was also awarded an ERC Advanced Grant, in 2011 for his project VarCity (Variation \& the City), was nominated ‘Distinguished Researcher’ by the IEEE Society of Computer Vision, in 2017, and received the 5-yearly excellence prize by the Flemish Fund for Scientific Research, in 2016. He received several other best paper awards as well. He is co-founder of several spin-offs. He has been involved in the organization of several, major conferences and as an associate editor for multiple, first-tier scientific journals.
\end{IEEEbiography}
\vskip -1\baselineskip plus -1fil

\begin{IEEEbiography}[{\includegraphics[width=1in,height=1.25in,clip,keepaspectratio]{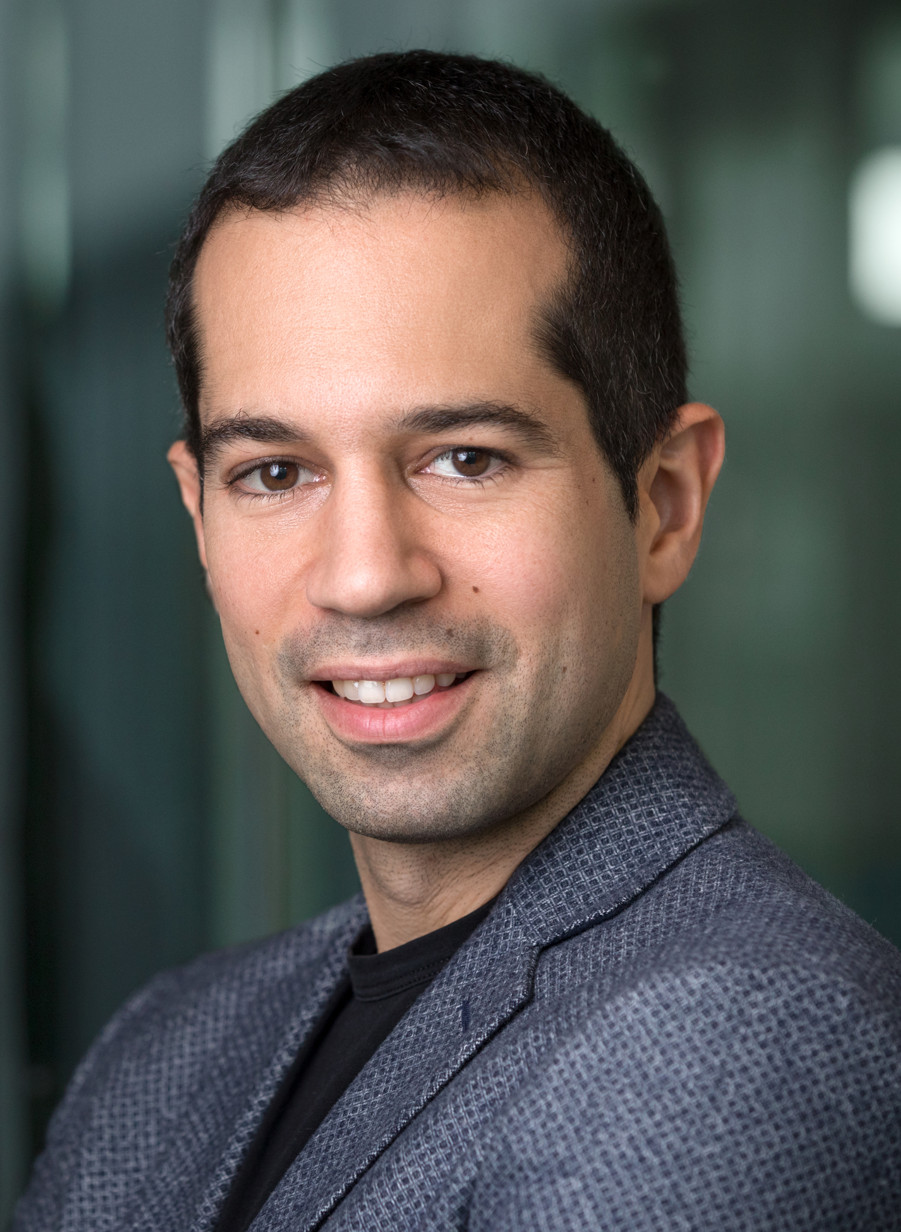}}]{Davide Scaramuzza} is a Professor of Robotics and Perception at the University of Zurich. He did his Ph.D. at ETH Zurich, a postdoc at the University of Pennsylvania, and was a visiting professor at Stanford University. His research focuses on autonomous, agile microdrone navigation using standard and event-based cameras. He pioneered autonomous, vision-based navigation of drones, which inspired the navigation algorithm of the NASA Mars helicopter and many drone companies. He contributed significantly to visual-inertial state estimation, vision-based agile navigation of microdrones, and low-latency, robust perception with event cameras, which were transferred to many products, from drones to automobiles, cameras, AR/VR headsets, and mobile devices. In 2022, his team demonstrated that an AI-controlled, vision-based drone could outperform the world champions of drone racing, a result that was published in Nature. He is a consultant for the United Nations on disaster response and disarmament. He has won many awards, including an IEEE Technical Field Award, the IEEE Robotics and Automation Society Early Career Award, a European Research Council Consolidator Grant, a Google Research Award, two NASA TechBrief Awards, and many paper awards. In 2015, he co-founded Zurich-Eye, today Meta Zurich, which developed the world-leading virtual-reality headset Meta Quest. In 2020, he co-founded SUIND, which builds autonomous drones for precision agriculture. Many aspects of his research have been featured in the media, such as The New York Times, The Economist, and Forbes.
\end{IEEEbiography}
\vskip -2.6\baselineskip plus -1fil

\vspace{11pt}

\vfill

\end{document}